\documentclass{article}

    \PassOptionsToPackage{numbers, compress}{natbib}

 \usepackage[main, final]{neurips_2026}

\usepackage[utf8]{inputenc} 
\usepackage[T1]{fontenc}    
\PassOptionsToPackage{dvipsnames}{xcolor}
\usepackage{hyperref}       
\usepackage{url}            
\usepackage{booktabs}       
\usepackage{amsfonts}       
\usepackage{nicefrac}       
\usepackage{microtype}      
\usepackage{xcolor}         

\def\thetitle{BA-T: An Iterative Transformer for\\[1mm] Two-View Bundle Adjustment}
\def\thetitlepdf{BA-T: An Iterative Transformer for Two-View Bundle Adjustment}

\title{\texorpdfstring{\thetitle}{\thetitlepdf}}
\newcommand{\ours}{BA-T\xspace}
\author{%
Ganlin Zhang$^{1,2}$ \qquad Weirong Chen$^{1,2}$ \qquad Daniel Cremers$^{1,2}$ \qquad Xi Wang$^{1,2,3,4}$\\
\vspace{0em}\\
$^{1}$ TU Munich \qquad $^{2}$ MCML \qquad $^{3}$ Ulm University \qquad $^{4}$ ETH Zurich
}

\usepackage{bm}
\usepackage{caption}
\usepackage{amssymb}
\usepackage{tabularx}
\usepackage{multirow} 
\usepackage{makecell} 
\usepackage{colortbl} 
\usepackage[Export]{adjustbox} 
\usepackage{arydshln} 

\usepackage{pifont} 
\usepackage{tikz} 
\usepackage{nicematrix}
\usepackage{enumitem}

\usepackage[ruled,vlined]{algorithm2e}
\usepackage{cleveref} 

\newcommand{\red}[1]{{\color{red}#1}}

\newcommand{\boldparagraph}[1]{\vspace{0pt}\noindent{\bf #1 \,}}

\definecolor{myPink}{rgb}{0.980, 0.502, 0.447} 
\definecolor{myOrange}{rgb}{1.000, 0.784, 0.486} 

\colorlet{colorFst}{Green!15}   
\colorlet{colorSnd}{SpringGreen!15}   

\colorlet{colorLow}{darkgray!60}    
\colorlet{colorLowc}{darkgray!98}    

\newcommand{\st}{\cellcolor{colorFst}\bf}   
\newcommand{\nd}{\cellcolor{colorSnd}\underline}      

\newcommand{\lo}{\color{colorLow}}          
\newcommand{\loc}{\color{colorLowc}}          
\def\with{\emph{w/}\space}
\def\without{\emph{w/o}\space}

\definecolor{darkgreen}{RGB}{0,130,0}
\newcommand{\green}[1]{\textcolor{darkgreen}{#1}}
\definecolor{customblue}{RGB}{111, 213, 245}   
\definecolor{custompink}{RGB}{232, 159, 247}   

\RequirePackage{xspace}
\makeatletter
\DeclareRobustCommand\onedot{\futurelet\@let@token\@onedot}
\def\@onedot{\ifx\@let@token.\else.\null\fi\xspace}
\def\eg{\emph{e.g}\onedot} 

\def\ie{\emph{i.e}\onedot}

\makeatother

\begin{document}

\maketitle
\begin{figure}[htbp]
    \centering
    \includegraphics[width=\linewidth]{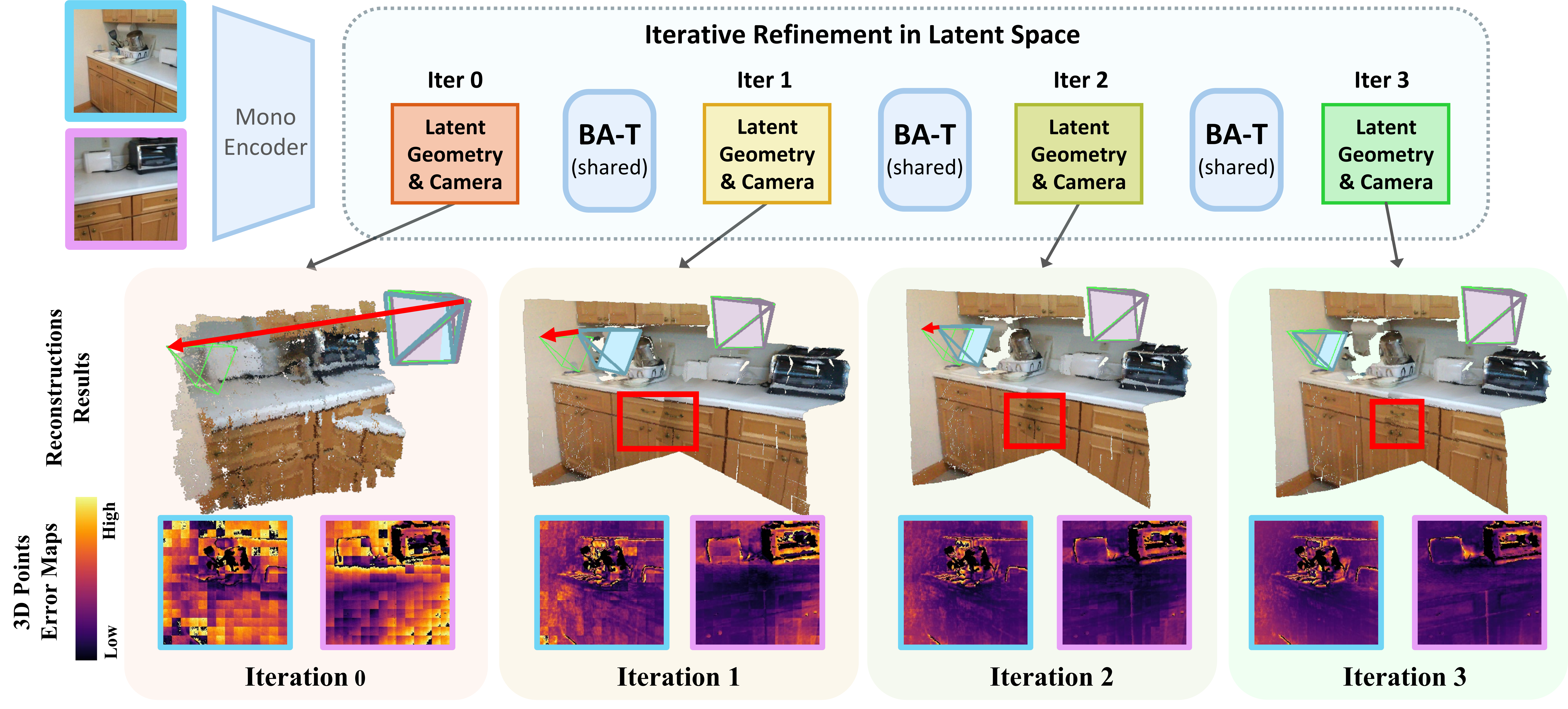}
\vspace{-0em}
\caption{
\textbf{Overview of \ours.} Given input images, \ours performs iterative updates on camera and local geometry tokens using a compact, reusable \ours layer in latent space. \textcolor{red}{$\longleftarrow$} indicates error between \green{GT poses} and estimated poses (\textcolor{customblue}{blue} and \textcolor{custompink}{pink}), which gradually decreases, and \textcolor{red}{red boxes} highlight regions progressively refined across iterations. The 3D point error maps visualize the per-view 3D point errors.
}
    \label{fig:teaser}
\end{figure}

\begin{abstract}

Feed-forward models for 3D reconstruction have achieved strong performance using deep cross-view attention to exchange information across images. However, these approaches often depend on heavy decoder stacks and lack a structured mechanism for geometry refinement, resulting in poor multi-view consistency.
We address this by drawing inspiration from classical bundle adjustment (BA), which can be viewed as an \emph{iterative information propagation process} between poses and local geometry. Inspired by BA, we propose \ours, an iterative Transformer that implements BA-style structured updates as a repeatable layer in implicit token space. Instead of relying on deep attention stacks, \ours refines predictions based on latent residual by a single lightweight layer.
Experiments demonstrate that \ours progressively improves pose and reconstruction accuracy across iterations, achieves stronger cross-view consistency than conventional decoders, and matches or surpasses substantially larger models while using only 16\% of their decoder parameters. \ours provides a compact, efficient, and structural alternative to depth-heavy attention, enabling accurate 3D reconstruction within a lightweight architecture.
Github repository: \url{https://github.com/zhangganlin/BA-T}.

\end{abstract}

\section{Introduction}

Estimating dense 3D structure and camera poses from multi-view images is fundamental in 3D computer vision, with applications in robotics, AR/VR, and autonomous driving.
To obtain dense reconstructions, recent feed-forward methods~\cite{dust3r,mast3r,wang2025vggt,keetha2026mapanything,wang2025pi3,depthanything3} directly regress dense 3D pointmaps (and camera poses) in a single forward pass. 
To enforce multi-view consistency, they use deep cross-view attention blocks as decoders to exchange information across images. While effective, these models rely on heavy decoder stacks to approximate geometric constraints and lack a mechanism to self-correct misalignments, limiting their robustness.

In contrast, classical bundle adjustment (BA)~\cite{triggs1999bundle,agarwal2010bundle,alismail2016photometric} refines camera poses and scene structure through iterative optimization, explicitly enforcing geometric consistency. However, its explicit geometric formulation makes it hard to embed in learning-based methods using implicit token representations.
Recently, there are learning-based 3D reconstruction systems~\cite{tang2018banet,teed2021droid,teed2023deep,wang2024vggsfm} trying to incorporate differentiable BA modules into their pipelines to improve reconstruction accuracy by adjusting predicted feature correspondences or optical flow. Nevertheless, these BA components still operate as external optimization stages: predictions are first converted into \textit{explicit geometric representations} (\eg, point clouds~\cite{tang2018banet,wang2024vggsfm} or feature tracks~\cite{teed2021droid,teed2023deep}), after which explicit second-order solvers are applied. Although differentiable, this design still performs iterative updates on explicit geometry, limiting its ability to leverage the expressive power of latent representations. Related iterative refinement strategies operating directly on latent representations have been explored for 3DGS~\cite{xu2025resplat}, but only for novel view synthesis rather than geometric reconstruction.


This reveals a key gap: feed-forward models leverage powerful latent token spaces but lack structured geometric refinement, while classical optimization is principled and iterative yet relies on explicit geometry, limiting integration with expressive latent representations.
To bridge this gap, we draw inspiration from the \emph{structured} iterative refinement of BA, where pose and geometry structure are progressively updated through repeated information propagation. Motivated by this, we propose \textbf{\ours} (\cref{fig:teaser}), a BA-inspired iterative Transformer that mimics BA-style information flow through repeatable learnable updates directly in implicit token space.

\ours replaces deep cross-view attention stacks with a single structured iterative refinement layer, enabling more compact and efficient updates of camera poses and 3D geometry than vanilla cross-attention blocks.
By combining latent 3D representations, feed-forward efficiency, and BA-inspired self-consistency, it bridges end-to-end regression and iterative optimization. Experiments show that \ours progressively improves pose and reconstruction accuracy, achieves stronger multi-view consistency than conventional decoder stacks, and matches or surpasses two-view models with up to $6\times$ more decoder parameters~\cite{dust3r} while remaining competitive with much larger multi-view models at 5--6\% of their decoder budget, showing that structured iterative refinement can replace depth-heavy attention while remaining lightweight and accurate.

In summary, our main \textbf{contributions} are as follows:
\begin{itemize}[itemsep=0pt,topsep=2pt,leftmargin=10pt,label=$\bullet$] 
\item We propose \ours, a compact learnable, \emph{bundle-adjustment-inspired} iterative Transformer layer for structured pose and geometry refinement.
\item We refine relative camera poses and local geometry at the token level in an \emph{implicit latent space}, leveraging the strong expressiveness of latent tokens.
\item We achieve superior accuracy and multi-view consistency with a single \emph{reusable} layer, matching or surpassing two-view models with up to $6\times$ more decoder parameters and remaining competitive with much larger multi-view models, while being more lightweight and computationally efficient.
\end{itemize}

\section{Related Work}
\boldparagraph{3D Reconstruction with Classical Bundle Adjustment.}
Classical 3D reconstruction systems~\cite{slam-handbook} use bundle adjustment (BA) to jointly optimize camera poses and 3D structure through iterative refinement. Sparse feature-based methods~\cite{davison2007monoslam,schonberger2016structure,mur2015orb,pan2024glomap,liu2024robust} estimate camera trajectories and 3D points from keypoints, refining them by minimizing reprojection error via BA. In contrast, direct methods~\cite{engel2014lsd,engel2017direct,gao2018ldso,yang2018deep,yang2020d3vo} optimize pixel intensities instead of sparse correspondences, formulating reconstruction as photometric error minimization.
Dense extensions, such as DROID-SLAM~\cite{teed2021droid} and others~\cite{hagemann2023droidcalib,zhang2023hi,zhang2024glorie,sandstrom2025splat,zhang2024hi} integrate learned priors with differentiable bundle adjustment, achieving strong geometric consistency through iterative optimization that explicitly couples pose and structure variables. However, these approaches still rely on solver-based explicit updates and require constructing and inverting large second-order Hessian matrices during inference. Such computational overhead makes seamless integration with modern large-scale feed-forward architectures challenging. Moreover, their dependence on optimization dynamics can limit robustness in texture-less or low co-visibility scenarios, where feed-forward models may instead benefit from large-scale data-driven priors.

\boldparagraph{3D Reconstruction with Feed-Forward Models.}
Recent learning-based methods perform geometry and pose estimation in a single forward pass. DUSt3R~\cite{dust3r} perform two-view reconstruction by regressing pointmaps in a shared global frame, while sequential extensions~\cite{wang2025spann3r,cabon2025must3r,wang2025cut3r} propagate reconstructions across multiple frames. Two-view follow-ups~\cite{zhang2025vistaslam,qian2026flow4r} predict local, view-dependent geometry for greater downstream flexibility. More recent models, including VGGT~\cite{wang2025vggt} and others~\cite{keetha2026mapanything,wang2025pi3,depthanything3}, are trained on multi-view data to generalize across more views, and SLAM-Former~\cite{slam-former} supports mixed inputs of reconstructed and new frames to enhance consistency.
A common characteristic of these feed-forward methods is the heavy use of cross-view attention blocks as geometric decoders for cross-view information exchange. Geometric consistency is enforced implicitly through deep stacks of attention layers, requiring increasingly large and computationally intensive models to approximate the constraints that classical optimization enforces explicitly. Despite their depth and capacity, these models lack an iterative correction mechanism at inference time: once a forward pass is completed, prediction errors cannot be systematically refined.

In contrast, our proposed \ours replaces heavy cross-view attention stacks with structured, repeatable refinement steps inspired by bundle adjustment. By embedding BA-inspired information propagation into a compact architecture, our method combines efficient feed-forward prediction with iterative geometric refinement without explicit solver-based optimization.
\setlength{\abovedisplayskip}{8pt}
\setlength{\belowdisplayskip}{8pt}

\section{\ours}
\subsection{Preliminary: Information Propagation of Bundle Adjustment}
Bundle Adjustment (BA) jointly optimizes scene geometry and camera parameters by minimizing residuals through iterative refinement. In this process, information circulates between camera estimates and geometry estimates through repeated updates. Many BA variants exist, depending on the choice of residuals and the representation of geometry.
We begin by presenting a general formulation of BA for the simplest two-view case. Given two images $\bm I_a$ and $\bm I_b$, BA aims to optimize their local geometry $\bm g_a$, $\bm g_b$ and their relative camera pose $c$. Consider a pair of local regions (e.g., a pixel or a patch) $p \in \bm I_a$ and $q \in \bm I_b$. The local geometry associated with $p$, denoted by $g_a^p\in \bm g_a$, is transformed into the local space of $\bm I_b$ via the space transform function $\pi_c^{a \rightarrow b}(\cdot)$. A property $f$ of the transformed local geometry is then compared with the corresponding measurement using a \emph{residual function} $r(\cdot, \cdot)$, which forms the core objective of bundle adjustment. The resulting optimization problem can be written as
{
\renewcommand\arraystretch{1.5}
\setlength{\arraycolsep}{3pt}
\begin{equation}
    {\hat c}, \bm{\hat g}_a, \bm {\hat g}_b = \operatorname*{arg\,min}_{{c}, \bm{g}_a,  \bm{g}_b} \left(
        \sum_{p\in \bm I_a, q\in\bm I_b} r\left(f(g_b^q), f\left(\pi_c^{a\rightarrow b} (g_a^p)\right)\right)
    \right),
    \label{eq:ba_general}
\end{equation}
}


To solve the optimization problem, the classical way uses a second-order Gauss-Newton solver~\cite{nocedal2006numerical}, taking use of the Jacobian matrix $\bm {J_c}$ and $\bm {J_g}$:
\begin{equation}
    \begin{bmatrix}
        \bm J_{\bm c}^T \bm J_{\bm c} & \bm J_{\bm c}^T \bm J_{\bm g} \\ 
        \bm J_{\bm g}^T \bm J_{\bm c} & \bm J_{\bm g}^T \bm J_{\bm g}
    \end{bmatrix}
    \begin{bmatrix}
        \Delta \bm c \\
        \Delta \bm g
    \end{bmatrix}
    = - 
    \begin{bmatrix}
        \bm J_{\bm c}^T \\
        \bm J_{\bm g}^T
    \end{bmatrix}\bm r 
    \xrightarrow{\text{rewrite as}}
    \begin{bmatrix}
        \bm B  & \bm E \\ 
        \bm E^T & \bm C
    \end{bmatrix}
    \begin{bmatrix}
        \Delta \bm c \\
        \Delta \bm g
    \end{bmatrix}
    = 
    \begin{bmatrix}
        \bm v \\
        \bm w
    \end{bmatrix}
    \label{eq:gaussnewton},
\end{equation}
and the update $\Delta \bm c$ and $\Delta \bm g$ can be solved by Schur Complement~\cite{hartley2003multiple}:
\begin{align}
    \Delta \bm c = 
    \begin{bmatrix}
        \bm B - \bm E\bm C^{-1}\bm E^T
    \end{bmatrix}^{-1} (\bm v -\bm E \bm C^{-1}\bm w), 
    \quad
    \Delta \bm g = \bm C^{-1}(\bm w -\bm E^{T} \Delta \bm c).
    \label{eq:delta_update_schur}
\end{align}
These equations summarize the information flow in BA: residuals are first computed, the camera update aggregates the coupled residual information through the Schur complement, and the resulting camera change is then used to refine the geometry. 

\begin{figure}[t]
    \centering
    \includegraphics[width=\linewidth]{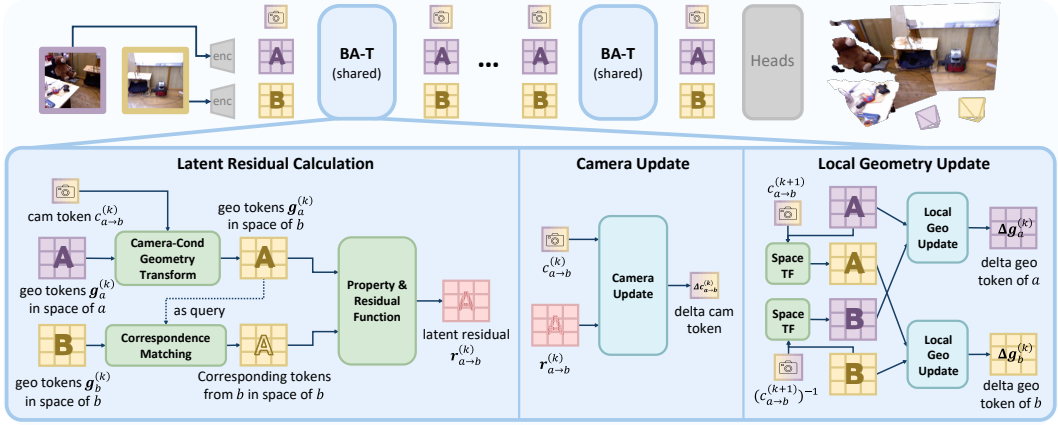}
\caption{\textbf{Overview of the \ours pipeline.} 
\ours takes camera tokens (from learnable initialization) and local geometry tokens (from the image encoder) as input and refines them iteratively. 
At each step, it performs a BA-inspired implicit refinement step by transforming geometry tokens across camera spaces, matching correspondences, and computing latent residuals. 
Camera tokens and per-view geometry tokens are refined via the Camera Update and Local Geometry Update respectively. }

\label{fig:pipeline}

\end{figure}

\subsection{Method Overview: BA-inspired Transformer for Implicit Refinement}
\boldparagraph{Notation.} We use the following typographic convention: for any token-valued quantity $x$, we write $x$ for a single token, $\bm x$ for a set of tokens, and $x^{(k)}$ for its value at refinement iteration $k$.

\boldparagraph{BA-T.}
Classical BA explicitly constructs and solves the normal equations (\cref{eq:gaussnewton}) via iterative second-order optimization, providing strong geometric guarantees but relying on explicit geometry, hand-crafted residuals, Jacobians, and large matrix inversions, making it difficult to integrate into modern end-to-end learning frameworks or scale to latent implicit representations.

The update equations~\cref{eq:delta_update_schur} already reveal the key pattern we want to preserve: information flows back and forth between camera and geometry through repeated refinement. Rather than explicitly forming and solving linear systems, we take this \emph{iterative information propagation} of BA as a design cue for a learnable refinement layer between camera parameters and local geometry. This dependency structure can be seen as a global information flow, where updates to one set of variables propagate and influence the other, motivating a learnable mechanism that emulates this interaction pattern.

We model this pattern of information propagation with a learnable layer, \textbf{\ours} (\cref{fig:pipeline}), treating camera and geometry parameters as latent tokens, and use attention layers to learn the interaction patterns suggested by classical BA. \ours is organized around three functional roles that implement this information flow: (1) cross-view correspondence estimation and latent residual computation, (2) camera update, and (3) geometry update. This BA-inspired  architecture realizes the update flow through structured token interactions that directly predict parameter updates. Thanks to the structured updates, \ours also outperforms naive iterative cross-attention stacks, as shown in \cref{sec:exp_iter}.

Formally, let $c_{a \rightarrow b} \in \mathbb{R}^{1 \times D}$ denote the camera token for an image pair $(\bm I_a, \bm I_b)$, where $D$ is the token dimension. The camera token encodes relative pose from view $a$ to view $b$. Let $\bm g_a, \bm g_b \in \mathbb{R}^{N \times D}$ denote the local geometry tokens for images $\bm I_a$ and $\bm I_b$, with $N$ tokens per image. The initial camera token $c^{(0)}_{a \rightarrow b}$ is a learnable parameter shared across all pairs, while the initial geometry tokens $\bm g_a^{(0)}$ and $\bm g_b^{(0)}$ are obtained from image features extracted by an image encoder.

\ours, denoted as  
$\mathcal{F}_{\theta}$,
takes the set of camera and geometry tokens as input, computes latent residuals in the implicit token space, and updates camera and geometry tokens accordingly. Specifically, given the token set $\mathcal{T}^{(k)}$ from iteration $k$, $\mathcal{F}_{\theta}(\cdot)$ produces updated tokens,
\begin{equation}
\mathcal{T}^{(k)} = \left\{ c^{(k)}_{a\rightarrow b}, \bm g_a^{(k)}, \bm g_b^{(k)} \right\},
\qquad
\Delta\mathcal{T}^{(k)} = \mathcal{F}_{\theta} \left(\mathcal{T}^{(k)}\right), \qquad  
\mathcal{T}^{(k+1)} = \mathcal{T}^{(k)} + \Delta\mathcal{T}^{(k)}.
\end{equation}
This update process can be applied iteratively for a variable number of steps $k = 0, \dots, K$, where $K$ is not fixed and may differ between training and inference. Through repeated refinement, \ours progressively enforces global geometric consistency across views, following the same broad refinement spirit as bundle adjustment while remaining a learned latent model. 

\begin{figure*}[!t]
\centering
\includegraphics[width=0.9\linewidth]{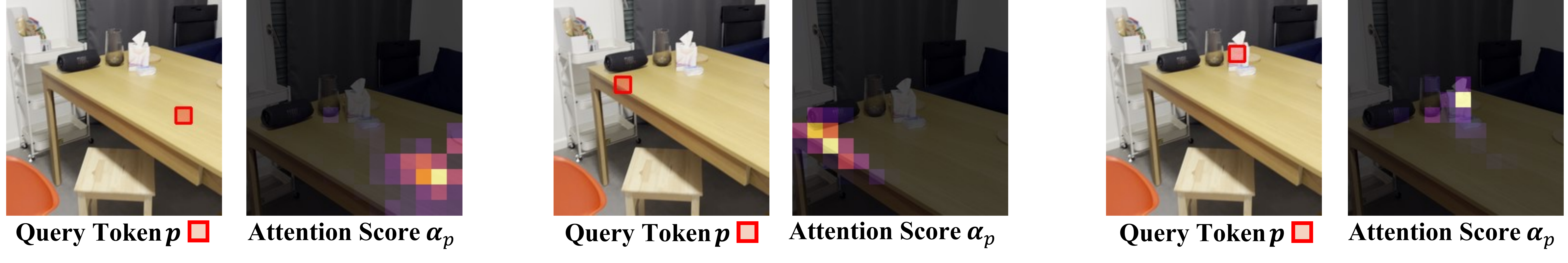}
\caption{
\textbf{Token-level correspondence response.} Given a local geometry token from one view, its attention scores highlight responses in the other view. Correct responses are observed for both ambiguous (left, middle) and distinctive (right) regions.
}
\label{fig:correspondence}
\end{figure*}
\subsection{Implementation of General Functions in \ours}
\label{sec:method_func}
Inspired by the general form of BA, we design the following components to support iterative refinement and information exchange between camera and geometry tokens: a camera-conditioned geometry alignment function $\pi_c^{a\rightarrow b}$, a property function $f$, and a residual function $r$. Additionally, given a geometry token $g_a^{p}$ in view $a$, we must identify its corresponding geometry token $g_b^{q}$ in view $b$.

\boldparagraph{Camera-Conditioned Geometry Transform.}
Since the geometry tokens of the two views are defined in their respective local frames, they are not directly comparable.
We first focus on how to inject the camera token $c_{a\rightarrow b}$ into the geometry token $g_a^{p}$ so that they can be compared consistently in a shared space. Since the local coordinates are linked via the relative transformation (which is encoded in the camera token), it is natural to aggregate information from the camera token $c_{a\rightarrow b}$ into the geometry token $g_a^{p}$ of view $a$.
Inspired by the conditioning mechanism in DiT~\cite{peebles2023dit}, we implement this transformation using a condition block with an adaLN layer, which conditions the local geometry token $g_a^p$ on the camera token $c_{a\rightarrow b}$:
$\pi_c^{a\rightarrow b}(g_a^{p}) := \text{Condition}(g_a^{p}, c_{a\rightarrow b})$.

\boldparagraph{Token-level Correspondence Matching.}
Once the transformed geometry token is obtained, we match it with the corresponding geometry token $g_b^{q}$ in view $b$ to compute the residual. Since each token represents a local image patch, the projected region in the other view may not align exactly with a single token as in pixel-level matching, but can instead fall between tokens or span multiple tokens. Therefore, a flexible correspondence matching mechanism is required.
We adopt an attention-inspired correspondence method. As shown in \cref{fig:correspondence}, for a given transformed token $\pi_c^{a\rightarrow b}(g_a^{p})$, a cross-attention layer queries the corresponding tokens in view $b$. 
We can then obtain the attention score $\bm \alpha_p$, measuring the relevance of the query token $\pi_c^{a\rightarrow b}(g_a^{p})$ and every token in view $b$, and the corresponding token $g_b^{q}$ is obtained as the attention-weighted sum of all geometry tokens in view $b$,
\begin{equation}
    \bm \alpha_p = \operatorname{AttnScore}\left(\pi_c^{a\rightarrow b}(g_a^{p}), \, \bm {g_b}\right), \qquad g_b^{q} = \sum_{i<N}\alpha_{p}^i \cdot g_b^{i}.
\end{equation}
This soft correspondence mechanism enables \ours to model sub-token matches and occlusions that are difficult to capture with explicit classical BA.

\begin{figure*}[!t]
\centering
\includegraphics[width=0.9\linewidth]{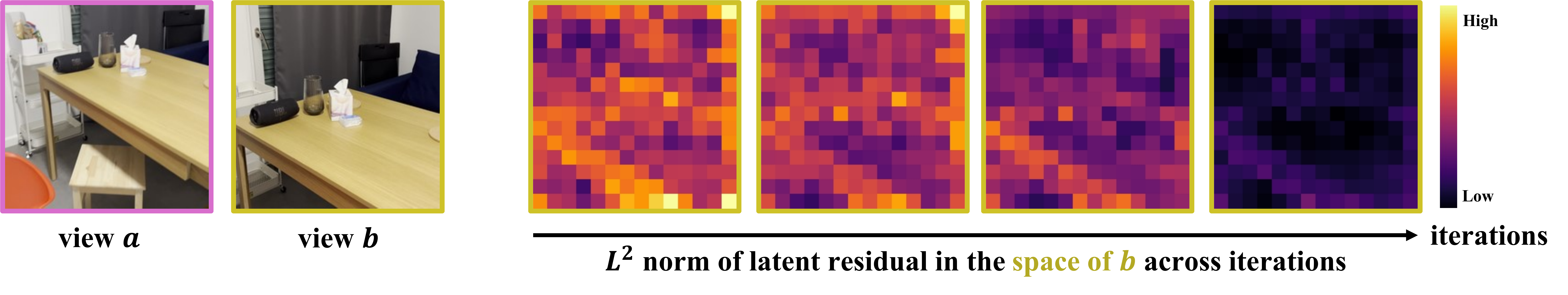}
\caption{
\textbf{Visualization of latent residuals.} Residuals are computed in the latent space of view $b$. Their magnitude decreases across refinement iterations, indicating increasingly accurate estimates.
}
\label{fig:residual}

\end{figure*}

\boldparagraph{Latent Property and Residual Function.}
For the property function $f(\cdot)$, we apply a linear projection to extract a suitable representation from each geometry token:
\begin{equation}
f(\pi_c^{a\rightarrow b}(g_a^{p})) = \operatorname{proj}\big(\pi_c^{a\rightarrow b}(g_a^{p})\big), \qquad
f(g_b^{q}) = \operatorname{proj}(g_b^{q}).
\end{equation}
The function $r(\cdot, \cdot)$ that calculates residual in latent space is implemented as an MLP applied to the difference between the properties of the two tokens:
\begin{equation}
r_{a\rightarrow b}\Big(f(g_b^{q}), f(\pi_c^{a\rightarrow b}(g_a^{p}))\Big) =
\operatorname{MLP}\Big(f(g_b^{q}) - f(\pi_c^{a\rightarrow b}(g_a^{p}))\Big).
\end{equation}
The resulting latent residual token encodes the discrepancy (\cref{fig:residual}) between the geometry in view $a$ (expressed in $b$'s space) and its corresponding region in view $b$, serving as a learned analogue of the classical residual (\eg, reprojection error).

\boldparagraph{Relative Camera Information Update}
\label{sec:method_cam_update}
Inspired by the information propagation pattern of classical BA in \cref{eq:delta_update_schur}, we update the camera token $c_{a\rightarrow b}$ first at each iteration in \ours.
In practice, the camera update $\Delta c_{a \rightarrow b}$ is predicted from the current camera estimate $c_{a \rightarrow b}$ and the residual tokens $\bm r_{a\rightarrow b}$, which already carry the geometry information needed for the next refinement step.
Each of the $N$ geometry tokens of view $a$ yields one residual token, so there are $N$ residual tokens $\bm r_{a\rightarrow b}$ per image pair.
Thus, we update the camera token by attending from the single camera token to all $N$ residual tokens. Specifically, we employ a one-to-many cross-attention mechanism, where the camera token serves as the query and the residual tokens act as the context to be aggregated,
\begin{equation}
\Delta c^{(k)}_{a\rightarrow b} =
\operatorname{CrossAttn}\left(
c^{(k)}_{a\rightarrow b},\;\bm r^{(k)}_{a\rightarrow b}
\right),
\end{equation}
where $\operatorname{CrossAttn}(\cdot, \cdot)$ denotes a cross-attention operator that aggregates information from multiple residual tokens conditioned on the current camera token. 
This one-to-many attention-based update implicitly captures the information flow of the camera update in Schur complement in classical BA.

\boldparagraph{Local Geometry Update}
After obtaining the camera update $\Delta c_{a \rightarrow b}$, we proceed to update the local geometries $\bm g_a$ and $\bm g_b$. As the residual information $\bm r_{a\rightarrow b}$ has already been aggregated into the camera update $\Delta c_{a \rightarrow b}$, the geometry update depends on the current estimates of local geometry $\bm g_a$ and $\bm g_b$, and the updated camera parameters $c_{a \rightarrow b}^{(k+1)} = c_{a \rightarrow b}^{(k)} + \Delta c_{a \rightarrow b}^{(k)}$.

To effectively exploit cross-view geometric consistency, we update the local geometry using cross-attention between the two views. Specifically, when updating the geometry of one view, we condition on both its own current geometry and the camera-conditioned geometry of the other view, so that both geometry are aligned in the same space,
\begin{align}
\Delta \bm g_a^{(k)} =
\operatorname{CrossAttn}\left(
\bm g_a^{(k)},
\bm \pi_{c^{(k+1)}}^{b \rightarrow a}\left(\bm g_b^{(k)}\right)
\right), \;
\Delta \bm g_b^{(k)} =
\operatorname{CrossAttn}\left(
\bm g_b^{(k)},
\bm \pi_{c^{(k+1)}}^{a \rightarrow b}\left(\bm g_a^{(k)}\right)
\right).
\end{align}

This symmetric cross-view update encourages consistent geometric refinement across views and follows the same information propagation pattern as the geometry update step in classical BA, where local structure is refined conditioned on the updated camera parameters.

Together, these camera and geometry update steps form a complete iterative refinement procedure in \ours, in which local geometry and relative camera parameters are progressively refined through repeated application of a single refinement layer.

\label{sec:method_geo_update}

\subsection{Implementation and Training}
\label{sec:method_training}
We instantiate \ours in a two-view reconstruction model that regresses local point maps and relative camera poses for an image pair, using a ViT~\cite{dosovitskiy2020image} encoder to capture per-image features and decoding the implicit token from interative \ours into per-view point maps and relative camera pose using regression heads (more details are provided in Appendix \cref{sec:implementation_details}).

\boldparagraph{Loss functions.}
The supervision signals include point-level geometry, relative pose, and cross-view consistency constraints, following standard choices in prior works~\cite{dust3r,dong2025reloc3r,zhang2025vistaslam}.
The overall training objective is defined as
\begin{equation}
L = \lambda_\text{pmap} L_\text{pmap}
+ \lambda_\text{pose} L_\text{pose}
+ \lambda_\text{gc} L_\text{gc},
\end{equation}
where $L_\text{pmap}$ supervises the predicted local point maps, $L_\text{pose}$ penalizes errors in relative rotation and translation with an additional cycle-consistency constraint, and $L_\text{gc}$ enforces geometric consistency between point maps across views under the predicted relative transformation. The weighting factors $\lambda_\text{pmap}$, $\lambda_\text{pose}$, and $\lambda_\text{gc}$ balance the three terms.

\boldparagraph{Iterative Supervision.}
Since \ours refines camera and geometry tokens over multiple iterations using the same Transformer layer, we apply supervision at the output of each iteration. Let $L^{(k)}$ denote the loss at iteration $k$, and let $K$ be the total number of iterations. To encourage progressive refinement, later iterations are given higher weights, while earlier iterations contribute less, This design ensures the network learns to gradually refine predictions rather than only focusing on the final output.
\vspace{-0.5em}
\begin{equation}
L_\text{total} = \sum_{k=1}^{K} \alpha_\lambda^{K-k} L^{(k)}, \quad( 0<\alpha_\lambda<1)
\label{eq:iterative_supervision}
\end{equation}

\section{Experiments}
We present the evaluation of \ours in three parts. First, we compare it with recent feed-forward 3D reconstruction models with deep stacked decoders to assess pose estimation and geometry reconstruction results. Next, we analyze the performance of iterative refinements. Finally, we ablate the components of \ours and discuss its potentials and limitations.
Best results are highlighted as \colorbox{colorFst}{\bf first} and \colorbox{colorSnd}{\underline{second}}.

\subsection{Evaluation of Pose Estimation and 3D Reconstruction}
In \cref{tab:pose_recon}, \ours (with $K=4$ iterations for inference) is compared with recent feedforward 3D reconstruction models trained with multi-view data~\cite{wang2025vggt,keetha2026mapanything,depthanything3} and with two-view data~\cite{dust3r,mast3r,zhang2025vistaslam}. 
Since these models employ much larger architectures than \ours, we train an additional variant of ViSTA~\cite{zhang2025vistaslam} with a decoder size comparable to \ours, denoted as ViSTA$^\dagger$, to enable a fair comparison. ViSTA$^\dagger$ shares the encoder, training data, training steps, loss and decoder budget with \ours and differs only in the decoder design; the full specification is given in Appendix \cref{sec:ablation_details}.


\boldparagraph{Metrics and Datasets.}
We evaluate both pose estimation and geometry reconstruction. For relative pose, we report AUC@$5^\circ$/$10^\circ$/$20^\circ$, computed as the area under the pose accuracy curve with thresholds $5^\circ/10^\circ/20^\circ$, where the pose error is the maximum of rotation and translation angular errors. For geometry, we report Chamfer distance, accuracy, completeness, and depth threshold accuracy $\delta_{1.25}$, defined as the percentage of pixels satisfying $\max(y_{\text{pred}}/y_{\text{gt}}, y_{\text{gt}}/y_{\text{pred}}) < 1.25$, capturing local geometric quality. Evaluation uses 3,600 and 1,109 co-visible image pairs from real world datasets 7Scenes~\cite{shotton2013-7scenes} and BundleFusion~\cite{dai2017bundlefusion}, which provide ground-truth camera poses and 3D geometry.

\begin{table}[t]
\caption{\textbf{Relative pose and geometry evaluation on 7Scenes~\cite{shotton2013-7scenes} and BundleFusion~\cite{dai2017bundlefusion}.} \ours($K=4$) achieves best  pose and competitive geometry performance among two-view methods, while maintaining a significantly smaller decoder size.}
\vspace{0.5em}
\centering
\resizebox{\textwidth}{!}{%
\setlength{\tabcolsep}{1pt}
\begin{tabular}{l ccc cccc ccc cccc c}
\toprule
& \multicolumn{7}{c}{7Scenes} 
& \multicolumn{7}{c}{BundleFusion} 
& \multirow{3}{*}{\makecell[c]{Decoder \\ Param \\ Size \#$\downarrow$}} \\

\cmidrule(lr){2-8} \cmidrule(lr){9-15}

& \multicolumn{3}{c}{Pose AUC} 
& \multicolumn{4}{c}{Geometry}
& \multicolumn{3}{c}{Pose AUC} 
& \multicolumn{4}{c}{Geometry} \\

\cmidrule(lr){2-4} \cmidrule(lr){5-8}
\cmidrule(lr){9-11} \cmidrule(lr){12-15}

Method 
& \makecell{@$5^\circ\uparrow$}  
& \makecell{@$10^\circ\uparrow$} 
& \makecell{@$20^\circ\uparrow$} 
& Chamf.$\downarrow$ 
& Acc.$\downarrow$ 
& Comp.$\downarrow$ 
& $\delta_{1.25}\uparrow$

& \makecell{@$5^\circ\uparrow$}  
& \makecell{@$10^\circ\uparrow$} 
& \makecell{@$20^\circ\uparrow$} 
& Chamf.$\downarrow$ 
& Acc.$\downarrow$ 
& Comp.$\downarrow$ 
& $\delta_{1.25}\uparrow$ \\

\hline
\multicolumn{16}{l}{\cellcolor[HTML]{EEEEEE}{\textit{Trained with Multi-view Data}}} \\ 

\lo VGGT~\cite{wang2025vggt}            & \lo 0.19 & \lo 0.39 & \lo 0.60 & \lo 0.10 & \lo 0.10 & \lo 0.07
& \lo 0.97 & \lo 0.18 & \lo 0.36 & \lo 0.56 & \lo\bf 0.06 & \lo\bf 0.07 & \lo\bf 0.04 & \lo\bf 0.99& \lo 605M \\
\lo MA~\cite{keetha2026mapanything}     & \lo 0.09 & \lo 0.23 & \lo 0.44 & \lo 0.10 & \lo 0.12 & \lo 0.07 & \lo \textbf{0.98} & \lo 0.03 & \lo 0.11 & \lo 0.27 & \lo 0.15 & \lo 0.19 & \lo 0.09 & \lo 0.98 & \lo 171M \\
\lo DA3~\cite{depthanything3}             & \lo \textbf{0.22} & \lo \textbf{0.43} & \lo \textbf{0.64} & \lo \textbf{0.08} & \lo \textbf{0.09} & \lo \textbf{0.06} & \lo \textbf{0.98} & \lo \bf 0.20 & \lo\bf 0.37 & \lo\bf 0.57 & \lo\bf 0.06 & \lo\bf 0.07 & \lo\bf 0.04 & \lo\bf 0.99& \lo 765M \\

\hline
\multicolumn{16}{l}{\cellcolor[HTML]{EEEEEE}{\textit{Trained with Two-view Data}}}\\

DUSt3R~\cite{dust3r}          & 0.09 & 0.22 & 0.42 & 0.13 & 0.13 & 0.12 & 0.81 & \st0.16 & \nd{0.33} &\nd{0.52} & 0.25 & 0.26 & 0.22 & 0.58 & 227M \\
MASt3R~\cite{mast3r}          & \nd{0.15} & 0.32 & 0.52 & 0.16 & 0.16 & 0.11 & 0.79 & \st0.16 & 0.32 & 0.50 & 0.24 & 0.26 & 0.22 & 0.58 & 227M \\
ViSTA~\cite{zhang2025vistaslam}           & 0.13 & \nd{0.33} & \nd{0.56} & \st0.10 & \st0.10 & \st0.10 & \st0.98 & 0.15 & 0.31 & 0.51 &\st 0.06 & \st0.06 & \st0.05 & \st0.99 & \nd{113M} \\
ViSTA$^\dagger$~\cite{zhang2025vistaslam}     & 0.01 & 0.06 & 0.17 & 0.15 & 0.12 & 0.15 & 0.94 & 0.02 & 0.07 & 0.17 & 0.09 & 0.09 & 0.09 & 0.98 & \st 38M \\
\textbf{\ours}  & \st0.16 & \st0.37 & \st0.59 & \nd{0.11} & \st0.10 & \nd{0.11} & \st0.98 & \st0.16 & \st0.34 & \st0.54 &\nd{0.07} &\nd{0.07} &\nd {0.06} & \st0.99 & \st38M \\
\bottomrule
\end{tabular}
}
\label{tab:pose_recon}


\end{table}

\boldparagraph{Results.} 
As shown in \cref{tab:pose_recon}, \ours achieves the best overall pose estimation among two-view models. 
It even delivers competitive results compared to models trained on multi-view data, despite having a significantly lighter structure (5\% of their size).
For geometry, \ours performs on par with substantially larger models.
Compared to a model with similar decoder size, \ie, ViSTA$^\dagger$~\cite{zhang2025vistaslam}, \ours consistently outperforms in both pose and geometry metrics, demonstrating the effectiveness of structured iterative refinement. 
Notably, it achieves this with a compact reusable layer (38M parameters), much smaller than prior methods (113M–765M), balancing accuracy and efficiency (running speed and memory footprint are reported in the Appendix \cref{tab:mem_time}).

\subsection{Performance under Iterative Refinements}
\label{sec:exp_iter}

\begin{table*}[t]
\centering

\caption{(left) \textbf{Performance comparison on 7Scenes~\cite{shotton2013-7scenes} across iterative refinement steps.} Increasing the number of refinement iterations consistently improves pose and geometry accuracy. ViSTA$^\dagger$ \with \textit{iterative training} performs iterative update by conventional stacked decoder with comparable size. \ours outperforms it at all iterations, showing the effectiveness of the proposed structural design.}
\label{tab:iter}

\begin{minipage}[t]{0.78\textwidth}
\vspace{0pt}
\centering

\setlength{\tabcolsep}{2pt}

\resizebox{\linewidth}{!}{%
\renewcommand{\arraystretch}{1.2}
\begin{tabular}{l c ccc ccc cccc c}
\toprule
 &  
& \multicolumn{3}{c}{Rot. AUC (deg)} 
& \multicolumn{3}{c}{Trans. AUC (m)} 
& \multicolumn{4}{c}{Geometry} & \multirow{2}{*}{\makecell{Decoder \\Size \#}} \\
\cmidrule(lr){3-5} \cmidrule(lr){6-8} \cmidrule(lr){9-12}
\makecell{Method}& \makecell{Iter}
& @5°$\uparrow$ & @10°$\uparrow$ & @20°$\uparrow$
& @0.05$\uparrow$ & @0.10$\uparrow$ & @0.20$\uparrow$
& Corr.$\downarrow$ & Rel.$\downarrow$ & $\delta_{1.05}\uparrow$ & $\delta_{1.25}\uparrow$ \\
\midrule
\multirow{1}{*}{\makecell[l]{ViSTA$^\dagger$}}
& / & 0.426 & 0.662 & 0.823 & 0.104 & 0.272 & 0.510 & 0.052 & 0.079 & 0.380 & 0.940 & \textbf{38M}\\ 

\midrule
\multirow{4}{*}{\makecell[l]{ViSTA$^\dagger$\\ \with iterative \\ training}}
& 1 & 0.442 & 0.676 & 0.830 & 0.101 & 0.270 & 0.505 & 0.062 & 0.061 & 0.497 & 0.967 & \multirow{4}{*}{\textbf{38M}}\\
& 2 & 0.472 & 0.699 & 0.844 & 0.133 & 0.322 & 0.555 & 0.048 & 0.060 & 0.513 & 0.968 \\
& 3 & 0.485 & 0.708 & 0.848 & 0.146 & 0.341 & 0.575 & 0.045 & 0.060 & 0.515 & 0.968 \\
& 4 & \nd{0.491} & \nd{0.712} & \nd{0.851} & \nd{0.151} & \nd{0.346} & \nd{0.580} & \nd{0.045} & \nd{0.060} & \nd{0.516} & \nd{0.968} \\

\midrule

\multirow{4}{*}{\textbf{\ours}}
& 1 & 0.650 & 0.819 & 0.908 & 0.322 & 0.555 & 0.749 & 0.046 & 0.051 & 0.569 & 0.972 & \multirow{4}{*}{\textbf{38M}}\\
& 2 & 0.671 & 0.830 & 0.914 & 0.401 & 0.624 & 0.794 & 0.032 & 0.050 & 0.585 & 0.975 \\
& 3 & 0.679 & 0.834 & 0.916 & 0.421 & 0.642 & 0.805 & 0.028 & 0.049 & 0.594 & 0.976 \\
& 4 & \st 0.683 & \st 0.836 & \st 0.917 & \st 0.422 & \st 0.643 & \st 0.805 & \st 0.026 & \st 0.048 & \st 0.601 & \st 0.976 \\
\bottomrule
\end{tabular}
}

\end{minipage}
\hfill
\begin{minipage}[t]{0.18\textwidth}
\vspace{-4pt}
\centering

\includegraphics[width=\linewidth]{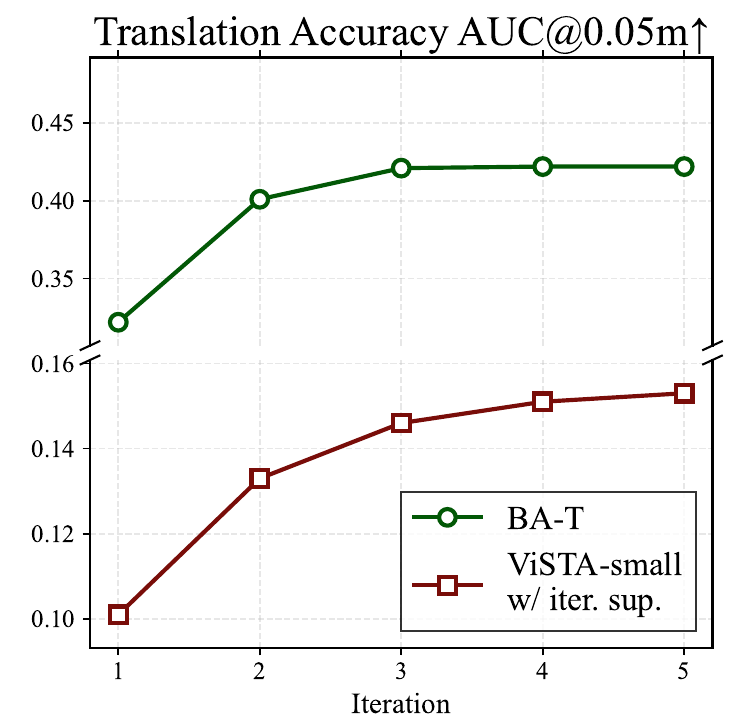}\\
\vspace{-4pt}
\includegraphics[width=\linewidth]{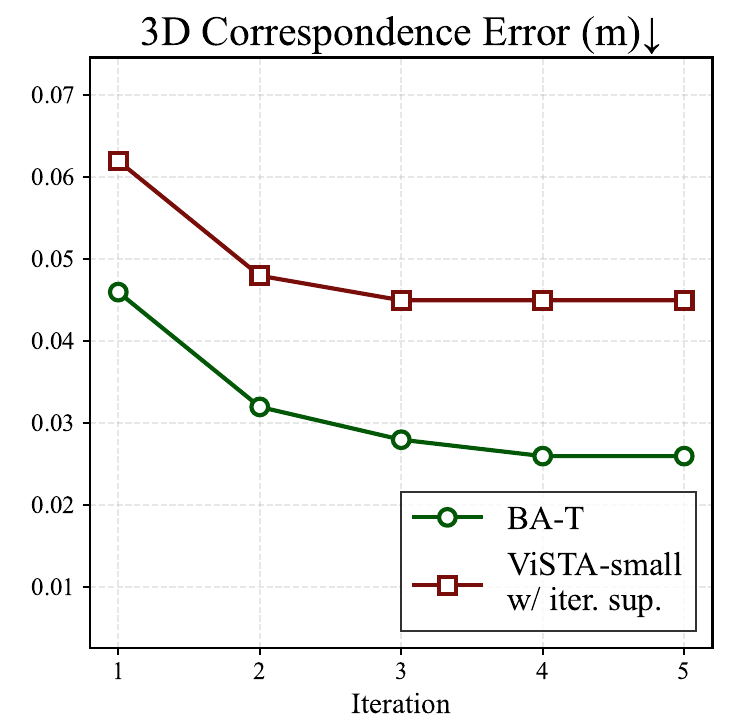}
\end{minipage}

\vspace{-0.6em}
\captionof{figure}{(right) \textbf{Iterative refinement behavior.}
\ours (\green{green curve}) consistently outperforms ViSTA$^\dagger$ \with iterative training (\textcolor{red}{red curve}) and converges within 3 $\sim$ 4 refinement steps.}
\label{fig:iter}
\end{table*}

\boldparagraph{Setup and Metrics.}
To evaluate the effectiveness of iterative refinement in \ours, we analyze performance across multiple update steps. For comparison, we also train a variant of ViSTA$^\dagger$~\cite{zhang2025vistaslam} using the same training settings as \ours. This variant, denoted as ViSTA$^\dagger$ \with \textit{iterative training}, employs iterative training in the same manner as \ours.
The only difference is that this baseline performs iteration with a conventional cross-view attention stacked decoder design, whereas \ours adopts a BA-inspired structured update mechanism.
\begin{figure*}[!t]
\centering
\includegraphics[width=\linewidth]{assets/bat_vis.pdf}
\caption{(left) \textbf{Qualitative reconstruction results.} We visualize reconstructed geometry and 3D points (in local frame) error maps at iteration~1 and iteration~4. \red{Red boxes} highlight noticeable misalignments in the $1_\text{st}$ iteration, which are corrected in the $4_\text{th}$ iteration, as indicated by \green{green boxes}. This demonstrates the effectiveness of the iterative refinement in \ours for improving local geometric consistency. 
(right) \textbf{Pose estimation visualization.} Estimated relative poses across different refinement iterations are shown. \ours progressively refines the pose estimates over iterations and remains robust even in low-visibility cases (second row).}
\label{fig:vis}


\end{figure*}
We report rotation (in degrees) and translation (in meters) AUCs, along with geometry metrics:
relative depth error (Rel.) and depth threshold accuracies $\delta_{1.05}$, $\delta_{1.25}$, to evaluate local geometry quality; 
3D correspondence error (Corr.), which is the Euclidean distance between corresponding 3D points from two images, and is used to assess the consistency of the local 3D reconstructions across the two views, with ground-truth correspondences obtained from GT depths and camera poses.

\boldparagraph{Quantitative Results.}
As shown in \cref{tab:iter}, iterative refinement in \ours steadily improves both pose and geometry. Both pose and geometry metrics get better across iterations, reflecting consistent gains in reconstruction. Most improvements occur within the first 3--4 iterations (\cref{fig:iter}), after which performance converges. For comparison, ViSTA$^\dagger$~\cite{zhang2025vistaslam} reduces 3D correspondence error by only 27\%, whereas \ours achieves 44\%, showing that our BA-inspired design more effectively enforces reconstruction consistency. Results demonstrate that structured iterative updates in \ours efficiently refine camera poses and 3D geometry, enabling high-accuracy reconstruction with a compact layer.
 
\boldparagraph{Qualitative Results.}
In \cref{fig:vis} (left), we show reconstruction results after 1 and 4 iterations, along with 3D points error maps, for scenes from 7Scenes~\cite{shotton2013-7scenes}, BundleFusion~\cite{dai2017bundlefusion}, and TUM-RGBD~\cite{sturm12tumrgbd}. After the first iteration, local geometry roughly captures the scene, but cross-view alignment is still inaccurate, producing misaligned planes and edges. Subsequent iterations progressively correct these discrepancies, yielding better-aligned structures, sharper details, demonstrating improved consistency through structured iterative updates.
In \cref{fig:vis} (right), it shows geometry and relative camera poses from iteration 1 to 3. Estimated poses become increasingly accurate with each iteration. Notably, in cases with very limited visual overlap (\eg, only a small co-visible region), \ours still gradually refines camera poses, demonstrating its robustness under challenging low-overlap conditions.

\begin{table*}[t]
\centering
\caption{
\textbf{Ablation study of \ours on 7Scenes~\cite{shotton2013-7scenes}.}
\emph{Every} row uses a \textbf{38M} decoder and shares the same encoder, training data, training steps and loss, so all differences are attributable to the decoder design alone.
Rows (a)--(b) use a conventional stacked cross/self-attention decoder with \emph{no} latent residual; rows (h) and (g) build up the BA-inspired structured update on top of (b); rows (c)--(e) remove one component from the full model.
The dominant factor is the latent residual itself: step (b)$\rightarrow$(h) alone recovers most of the gain over an identically-trained conventional decoder.
}
\setlength{\tabcolsep}{2pt}
\resizebox{\textwidth}{!}{%
\begin{NiceTabular}{cl|ccc ccc | cccc}
\toprule
& & \multicolumn{3}{c}{Rot. AUC (deg)}
& \multicolumn{3}{c}{Trans. AUC (m)}
& \multicolumn{4}{c}{Geometry Reconstruction} \\

\cmidrule(lr){3-5} \cmidrule(lr){6-8} \cmidrule(lr){9-12}
& Setting
& @5$^\circ\uparrow$ & @10$^\circ\uparrow$ & @20$^\circ\uparrow$
& @0.05$\uparrow$ & @0.10$\uparrow$ & @0.20$\uparrow$
& Corr.$\downarrow$
& Rel.$\downarrow$
& $\delta_{1.05}\uparrow$
& $\delta_{1.25}\uparrow$ \\
\midrule

\multicolumn{12}{l}{\cellcolor[HTML]{EEEEEE}{\textit{Conventional stacked decoder (no latent residual)}}} \\

(a) & ViSTA$^\dagger$ \textit{(no iteration)}
& 0.426 & 0.662 & 0.823
& 0.104 & 0.272 & 0.510
& 0.052 & 0.079 & 0.380 & 0.940 \\

(b) & ViSTA$^\dagger$ \with \textit{iterative training}
& 0.491 & 0.712 & 0.851
& 0.151 & 0.346 & 0.580
& 0.045 & 0.060 & 0.516 & 0.968 \\

\midrule
\multicolumn{12}{l}{\cellcolor[HTML]{EEEEEE}{\textit{Building up the BA-inspired structured update on top of (b)}}} \\

(h) & \; + latent residual \textit{(minimal \ours)}
& 0.637 & 0.810 & 0.904
& 0.296 & 0.534 & 0.735
& 0.039 & 0.051 & 0.565 & 0.971 \\

(g) & \; ++ geo. trans. \& corr. matching \textbf{(full \ours)}\, 
& \st 0.683 & \st 0.836 & \st 0.917
& \st 0.422 & \st 0.643 & \st 0.805
& \st 0.026 & \st 0.048 & \nd{0.601} & \st 0.976 \\

\midrule
\multicolumn{12}{l}{\cellcolor[HTML]{EEEEEE}{\textit{Removing one component from the full model}}} \\

(c) & \ours \without geometry transform
& 0.665 & 0.827 & 0.912
& 0.385 & 0.613 & 0.787
& \nd{0.027} & 0.050 & 0.599 & \nd{0.974} \\

(d) & \ours \without iterative supervision
& 0.667 & 0.827 & 0.912
& 0.404 & 0.625 & 0.793
& 0.036 & \nd{0.049} & \st 0.620 & \nd{0.974} \\

(e) & \ours \without correspondence matching
& \nd{0.677} & \nd{0.833} & \nd{0.916}
& \nd{0.412} & \nd{0.635} & \nd{0.802}
& \st 0.026 & \st 0.048 & 0.591 & \st 0.976 \\

\bottomrule
\end{NiceTabular}
}
\label{tab:ablation}
\end{table*}

\subsection{Ablation: Effectiveness of Components in \ours}
\label{sec:exp_ablation}
In \cref{tab:ablation}, we ablate \ours as a single ladder at a fixed 38M decoder budget. Crucially, the two ViSTA$^\dagger$ baselines are included as ablation rows: they share the encoder, training data, training steps and loss with \ours and differ only in the decoder design, so they separate the contribution of \emph{iteration} from that of the \emph{BA-inspired structure}. Detailed descriptions of each setting are given in Appendix \cref{sec:ablation_details}.

\boldparagraph{Iteration alone is not the contribution.}
Adding iterative training to a conventional stacked decoder, (a)$\rightarrow$(b), recovers only 25\% of the total rotation gain, 15\% of the translation gain and 27\% of the cross-view consistency gain. The remaining 75\% / 85\% / 73\% comes from replacing the stacked decoder with the BA-inspired structured update, (b)$\rightarrow$(g).

\boldparagraph{The latent residual is the dominant factor.}
To identify \emph{what} in the structured update produces that gap, row (h) adds only the simplest possible residual mechanism to (b), with every other \ours component disabled, so that the two views communicate through an explicit latent residual instead of through direct cross-attention. This single change already recovers 76\% of the rotation gain, 54\% of the translation gain and 75\% of the relative depth gain of (b)$\rightarrow$(g). Forming an explicit residual measurement inside the loop, even the crudest one, is therefore what drives the iterative refinement.

\boldparagraph{The remaining components improve the quality of the residual.}
The geometry transform and correspondence matching, (h)$\rightarrow$(g), make the residual more accurate and informative: they account for the majority of the cross-view consistency gain (68\% of the Corr. improvement) and for a further 46\% of the translation gain. This is consistent with the leave-one-out rows. Removing the geometry transform (c) forces the network to operate directly in the two different local coordinate frames of the input views, degrading both pose and geometry and confirming that cross-view alignment is essential for effective token updates. Removing correspondence matching (e) mainly costs relative pose while leaving per-view geometry largely unchanged, suggesting these tokens facilitate cross-view pose alignment more than per-view geometry.

\boldparagraph{Iterative supervision.}
In row (d) the decoder is still applied $K$ times, but only the output of the \emph{final} iteration is supervised during training, instead of supervising the decoded result of every refinement iteration as in \cref{eq:iterative_supervision}. This lowers rotation and translation accuracy and increases 3D correspondence error, indicating that supervising every iteration is what makes the refinement progressive. Overall, the full \ours achieves the strongest results, highlighting the importance of structured token interaction for accurate pose estimation and geometry.

\begin{figure*}[t]
\centering
\includegraphics[width=\linewidth]{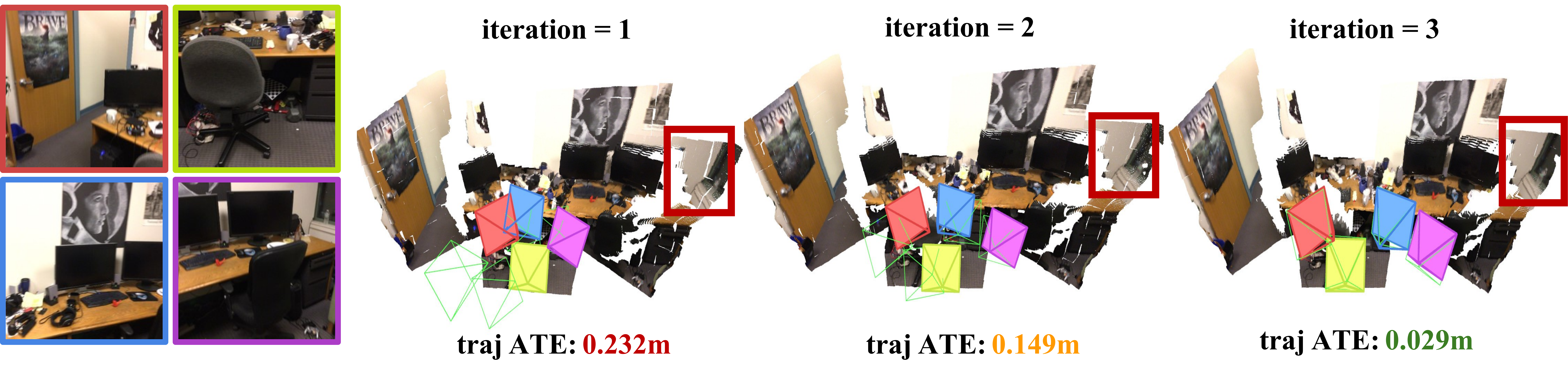}
\caption{
\textbf{Multiview reconstruction results.}
Estimated poses and reconstructed scenes are visualized across iterations for a 4-view input setup, demonstrating \ours's ability to handle multi-view settings. The green frustums indicate GT camera poses. The red boxes and decreasing trajectory ATE (\red{0.232m} $\rightarrow$ \green{0.029m}) highlights the effectiveness of iterative refinements.
}
\label{fig:multiview}
\end{figure*}

\subsection{Extension: \ours for Multiview Reconstruction}
\label{sec:exp_mv}
Thanks to the BA-inspired design, all cross-view interactions in \ours are performed in view's local space, making it possible to directly extend to multi-view inputs \emph{\textbf{without}} architectural changes. The most straightforward way for multiview extension is to allow each query view to attend to all other views, producing per-view pointmaps (details in Appendix). Visualization of a multiview example is shown in \cref{fig:multiview}. More importantly, the model refines all related views together, allowing information from one view to immediately help improve the others and strengthen overall consistency. This makes \ours a natural fit for multi-view reconstruction and allows it to scale beyond the two-view setting. 

While our design is not inherently limited to two views, the current implementation and training mainly focus on two-view reconstruction, broader multi-view configurations remain unexplored. Future work includes integrating \ours into multi-view reconstruction frameworks~\cite{wang2025vggt,depthanything3} to replace cross-view attention decoders and leverage large-scale data, potentially improving global consistency and reconstruction accuracy.
\section{Conclusion} 
We propose \ours, a Bundle-Adjustment-inspired iterative Transformer that jointly refines camera poses and 3D geometry via an implicit, structured, learnable BA-inspired layer. Unlike conventional feed-forward decoders that predict in a single pass, \ours enables coordinated, progressive refinement within a compact architecture. Experiments show it consistently improves pose and geometry estimates across iterations, achieves stronger consistency than standard stacked-attention decoders, and delivers high-accuracy pose estimation and 3D reconstruction efficiently.

\boldparagraph{Acknowledgment} This research was funded by the German Federal Ministry of Education and Research through
the ExperTeam4KI funding program for UDance (Grant
No. 16IS24064).


\newpage
\bibliographystyle{splncs04}
\bibliography{main}


\appendix
\clearpage

{
\centering
\Large
\textbf{\thetitle}\\
\vspace{0.5em}Appendix \\
\vspace{1.0em}
}
\section{Architecture and Training Details}
\label{sec:implementation_details}
\boldparagraph{Architecture Details.}
In \ours, tokens have dimension $D = 768$. In the Space Transform module, geometry tokens are transformed using an AdaLN-style conditional block~\cite{peebles2023dit}, where each token is normalized and modulated with shift and scale vectors predicted from the camera token (also 768-D). Token-level Correspondence Matching uses 8 attention heads, enabling tokens from different views to interact and establish geometric correspondences. The Latent Property and Residual Function applies linear projections to matched tokens, followed by a 2-layer MLP with expansion ratio 4 to produce normalized residuals that inform the camera update. The camera token is updated through cross-attention (12 heads) and a 2-layer feed-forward MLP with expansion ratio 4. Local geometry tokens are refined via cross-view attention (12 heads, 2-layer MLP with expansion ratio 4) and three intra-view self-attention layers (each with 12 heads and a 2-layer MLP with expansion ratio 4) to improve per-view feature consistency. Training uses $K = 5$ iterations for iterative supervision.

\boldparagraph{Training Details.}
We initialize the encoder with a pretrained ViT~\cite{zhang2025vistaslam}, while \ours and the regression heads are randomly initialized. All components are trained jointly.
Following the setting of prior work~\cite{zhang2025vistaslam}, \ours is trained on ScanNet~\cite{dai2017scannet}, ScanNet++~\cite{yeshwanthliu2023scannetpp}, ARKitScenes~\cite{dehghan2021arkitscenes}, CO3D~\cite{reizenstein21co3d}, ASE~\cite{avetisyan2024ase}, and Replica~\cite{replica19arxiv}, for 2 days on 8 NVIDIA H100 GPUs. Training uses AdamW~\cite{loshchilov2017adamw} with a learning rate of $1.5\times10^{-4}$, weight decay 0.01, and $\alpha_\lambda = 0.8$, $K = 5$. 

\section{Baseline and Ablation Details}
\label{sec:ablation_details}
This section specifies the ViSTA$^\dagger$ baselines used in \cref{tab:pose_recon}, \cref{tab:iter} and \cref{tab:ablation}, and describes each ablation setting of \cref{tab:ablation} in detail. Every model discussed here has a \textbf{38M} decoder, so that all reported differences are attributable to the decoder design alone.

\boldparagraph{ViSTA$^\dagger$ (row a).}
Identical in kind to the original ViSTA~\cite{zhang2025vistaslam} decoder: a stack of identical decoder layers, each with (i) self-attention applied to each view separately, followed by (ii) cross-attention in which each view's tokens act as queries and the other view's tokens as keys and values. The original ViSTA decoder has \textbf{12} such layers (113M); we use \textbf{3} so that the decoder parameter count (38M) matches \ours exactly.

\boldparagraph{ViSTA$^\dagger$ \with iterative training (row b).}
Identical to ViSTA$^\dagger$, but trained with the iterative strategy of \ours: the 3-layer decoder is applied $K$ times, the $K$ intermediate token sets are decoded by the same regression heads, and supervision is the weighted sum of \cref{eq:iterative_supervision}.

\boldparagraph{Controlled factors.}
Both baselines use the same pretrained ViSTA encoder as \ours, with decoder and regression heads randomly initialised; the same training datasets; the same number of training steps; and the same loss with the same loss weights. Since each \ours training step backpropagates $L_\text{total}$, a weighted sum over refinement outputs, we additionally multiply the non-iterative ViSTA$^\dagger$ loss by $\sum_{k=1}^{K}\alpha_\lambda^{K-k}$ so that the loss magnitude per backward pass is comparable across all three models. Without this control the comparison would be confounded by the effective learning rate.

\boldparagraph{Ablation settings.}
The remaining rows of \cref{tab:ablation} are defined as follows.
\begin{itemize}[itemsep=1pt,topsep=2pt,leftmargin=12pt,label=$\bullet$]
\item \textbf{(h) Minimal \ours}: the simplest possible BA-inspired decoder, obtained by adding only a latent residual to row (b). The geometry transform and the correspondence matching are both disabled, so the residual reduces to its crudest form: for each of the $N$ token positions we subtract the two views' tokens at the same 2D coordinate and pass the difference through an MLP, $r^{i}=\operatorname{MLP}(g_a^{i}-g_b^{i})$, in place of the correspondence and residual computation of \cref{sec:method_func}. The difference between (b) and (h) is therefore \emph{only} the medium of cross-view communication: (b) lets the two views' tokens cross-attend to each other directly, whereas (h) forms an explicit latent residual and conditions the update on it. Everything else (encoder, data, steps, loss, iterative training and supervision) is identical.
\item \textbf{(c) \without geometry transform}: all camera-conditioned geometry transform operations $\pi_c^{a\rightarrow b}(\cdot)$ are replaced by the identity.
\item \textbf{(d) \without iterative supervision}: the decoder is still applied $K$ times, but only the output of the final iteration is supervised, instead of the decoded result of every refinement iteration.
\item \textbf{(e) \without correspondence matching}: when computing the residual, we directly use the view $b$ token at the same 2D coordinate as the view $a$ token, instead of the attention-fused corresponding token.
\end{itemize}
Note that rows (c)--(e) are leave-one-out ablations measured from the full model, so they all still contain the BA-inspired residual mechanism and cannot on their own isolate what separates the iteratively-trained conventional decoder from \ours. Row (h) is included for exactly that purpose.

\section{Bidirectional Camera Representation}
In \ours, each two-view pair is represented by a single camera token that encodes the relative transformation. However, the local geometry update requires both directions of the relative transformation, and the camera-conditioned operator $\pi_{c^{(k+1)}}^{b \rightarrow a}(\cdot)$ still depends on the camera parameters $c_{b \rightarrow a}^{(k+1)}$.
To obtain the opposite direction of the transformation, we recover $c_{b \rightarrow a}$ by applying a lightweight MLP to $c_{a \rightarrow b}$. This MLP learns the inverse transformation in latent space, which avoids redundant camera tokens while preserving bidirectional geometric consistency.

\section{Solver-based BA Comparison}
\begin{figure*}[!ht]
\centering
\includegraphics[width=0.95\linewidth]{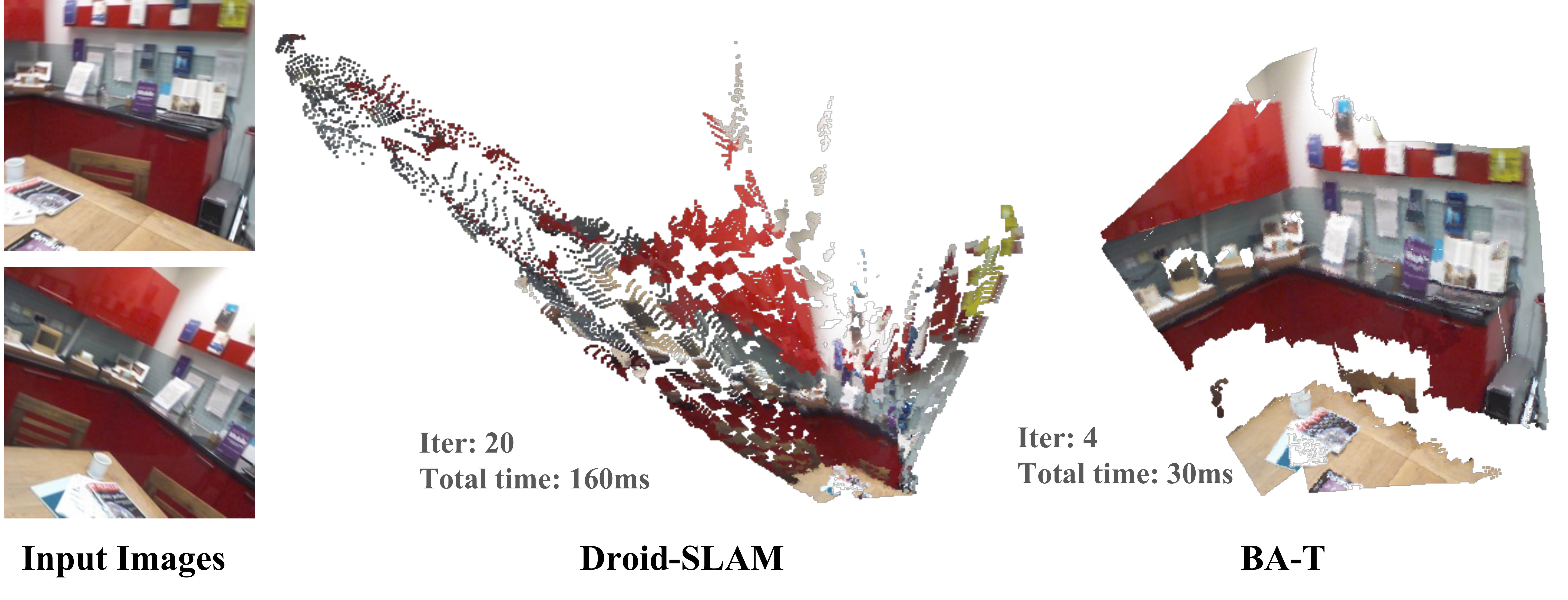}
\caption{
\textbf{Reconstruction comparison.} Both methods are initialized from \ours (iter = 0). Droid-SLAM uses ground-truth intrinsics, whereas \ours does not. \ours achieves stronger performance with fewer iterations and faster runtime, benefiting from its more expressive latent space.
}
\label{fig:droid}
\end{figure*}

We further compare \ours with Droid-SLAM, which relies on classical bundle adjustment (BA) using reprojection error as the residual and Gauss–Newton as the optimizer. For a fair comparison, both methods are initialized with the same local geometry and camera poses (taken from iteration 0 of \ours). Since Droid-SLAM requires camera calibration, we additionally provide the ground-truth intrinsic matrix.
Quantitative and qualitative results are reported in \cref{tab:droid} and \cref{fig:droid}, respectively. The results show that classical BA underperforms compared to \ours, particularly in geometry reconstruction. This is largely due to the highly non-convex optimization objective, which is difficult to solve reliably under limited co-visibility constraints (e.g., few-view settings). In contrast, \ours benefits from expressive latent representations and learned priors from training data, leading to improved robustness and accuracy.
Moreover, \ours is substantially more efficient, achieving approximately 5$\times$ faster inference while reducing memory consumption by about 45\%.

\begin{table}[ht]
\centering
\caption{\textbf{Quantitative comparison of two-view inputs on 7Scenes~\cite{shotton2013-7scenes}.}  }
\resizebox{\textwidth}{!}{%
\setlength{\tabcolsep}{6pt}
\renewcommand{\arraystretch}{1.1}
\begin{tabular}{lcccccc}
\toprule
Method 
& Rot AUC @5$^\circ$
& Trans AUC @0.05m
& Rel. 
& $\delta<1.25$ 
& Time (ms) 
& Peak Mem (GB) \\
\midrule
Droid-SLAM~\cite{teed2021droid} & 0.53 & 0.08 & 1.19 & 0.36 & 160.93 & 2.43 \\
\textbf{\ours}   & \bf{0.68} & \bf{0.42} & \bf{0.20} & \bf{0.98} & \bf 30.92 & \bf 1.35 \\
\bottomrule
\end{tabular}
}
\label{tab:droid}
\end{table}

\section{Running Time and Peak Memory}

We evaluate \ours against several multiview~\cite{wang2025vggt,depthanything3,keetha2026mapanything} and two-view~\cite{teed2021droid,mast3r,dust3r,zhang2025vistaslam} methods under the two-view setting, measuring decoder size, peak GPU memory, and inference time on the same input images with the same resolution.
As shown in \cref{tab:mem_time}, \ours achieves the most compact architecture with only 38M parameters, substantially smaller than all prior methods. It also requires the lowest peak memory (1.32 GB) and delivers faster inference, demonstrating a significant improvement in computational efficiency while maintaining competitive accuracy. These results highlight the effectiveness of BA-T’s compact and iterative design.

\begin{table}[htb]
\centering
\caption{\textbf{Model size and efficiency comparison under two-view inputs.} Decoder size is reported in millions of parameters (M), peak GPU memory in GB, and running time in milliseconds (ms). \ours achieves the most compact architecture and the highest computational efficiency among all compared methods.}
\setlength{\tabcolsep}{3pt}
\renewcommand{\arraystretch}{1.1}
\resizebox{\linewidth}{!}{
\begin{tabular}{lcccccccc}
\toprule
& Droid & VGGT & DA3 & MapAny & MASt3R & DUSt3R & ViSTA 
& \textbf{\ours}  \\
& {\cite{teed2021droid}}
& {\cite{wang2025vggt}} 
& {\cite{depthanything3}} 
& {\cite{keetha2026mapanything}} 
& {\cite{mast3r}} 
& {\cite{dust3r}} 
& {\cite{zhang2025vistaslam}} 
& ($k{=}1 / 2 / 3 / 4$)\\
\midrule
Running time $\downarrow$ (ms) & 160.93 & 128.71 & 78.14 & 41.13 & 93.25 & 66.47 & 37.96 
             & \loc 21.40 / \loc 24.65 / \loc 28.32 / \bf 30.92 \\
Decoder size $\downarrow$ (M) & / & 605 & 765 & 171 & 227 & 227 & 113 
             & \multicolumn{1}{c}{\bf 38} \\
Peak GPU mem $\downarrow$ (GB) & 2.43 & 4.93 & 6.52 & 3.23 & 2.72 & 2.02 & 1.76 
             & \multicolumn{1}{c}{\bf 1.32} \\
\bottomrule
\end{tabular}
}
\label{tab:mem_time}
\end{table}

\section{Discussion on Iteration and Refinement Behavior}

\subsection{Comparison with Unrolling and Recurrent Methods}
Although \ours is inspired by the iterative refinement spirit of bundle adjustment, it differs from prior unrolling and recurrent methods in both the representation it updates and the role played by each iteration.
BA-Net~\cite{tang2018banet}, RAFT~\cite{teed2020raft}, ReSplat~\cite{xu2025resplat}, and \ours all share an iterative refinement flavor, but the differences are substantial. BA-Net unrolls BA-style updates on explicit geometric variables, RAFT iteratively refines dense correspondence fields for optical flow, and ReSplat recurrently updates latent representations for Gaussian-splat novel-view synthesis; in all three cases, the update target and problem setting differ a lot from ours. The ablation with ViSTA$^\dagger$ using iterative training supports the same point: iterative training helps, but a large gap remains compared to \ours, indicating that iteration alone is not enough. By contrast, \ours performs the refinement entirely in latent space and is designed around the BA structural information propagation path, where camera and geometry tokens exchange information through a compact, reusable refinement layer. In other words, our method is not just iterative in the generic sense: it is specifically structured to propagate information between camera and geometry in a reconstruction setting, rather than unrolling an explicit solver or recurrently refining flow or rendering variables.

\subsection{Pose and Geometry Behavior in Experiments}
In our experiments, pose estimation improves more consistently than geometry reconstruction, while the geometry gains are more limited. This behavior follows from the update design: the residual tokens directly influence the camera update, so pose benefits from a more immediate refinement signal. Geometry, in contrast, is updated after the camera has been refined, so its improvement is more indirect.

A second factor is an \textbf{initialisation asymmetry}, which applies equally to \ours and to all baselines in our comparison. The camera token is initialised from a single shared learnable token, which we verified decodes to approximately the identity transform; it therefore starts far from the solution and the decoder has to produce essentially the whole pose, leaving considerable room to improve. The geometry tokens, by contrast, are initialised from the pretrained encoder, which already carries a strong monocular geometry prior and already yields a good per-view prediction. Most of the final geometry accuracy is thus in place before the decoder runs, so the decoder performs alignment and minor refinement rather than construction.

A third factor is \textbf{dimensionality}. Pose is a single 6-DoF quantity that is heavily overdetermined by many correspondences, so aggregating $N$ residuals into one update filters noise effectively and benefits pose disproportionately. Geometry is high-dimensional and per-token, so the same aggregated signal is structurally weaker for it.

Another factor is that we reconstruct geometry on a per-view basis. The final scene geometry is therefore formed by combining the local geometry predictions with the estimated camera poses, rather than by optimizing a single global geometry representation at once. As a result, the local geometry outputs may not look dramatically different even when the pose estimates improve substantially.

\boldparagraph{When geometry refinement helps and when it does not.}
Geometry refinement succeeds where the error is an \emph{alignment} error between two locally plausible reconstructions, which is precisely what the latent residual can express. It contributes least in two regimes: where the encoder's monocular prior is already accurate, leaving little to correct, and where co-visibility is almost absent, so that no meaningful residual can be formed in the first place.

This is why the quantitative gains are often clearer for pose than for geometry. The refinement process mainly strengthens the camera-geometry interaction, and the benefit to reconstruction is accumulated through the updated poses and the per-view local reconstructions.

\begin{figure*}[t]
\centering
\includegraphics[width=0.9\linewidth]{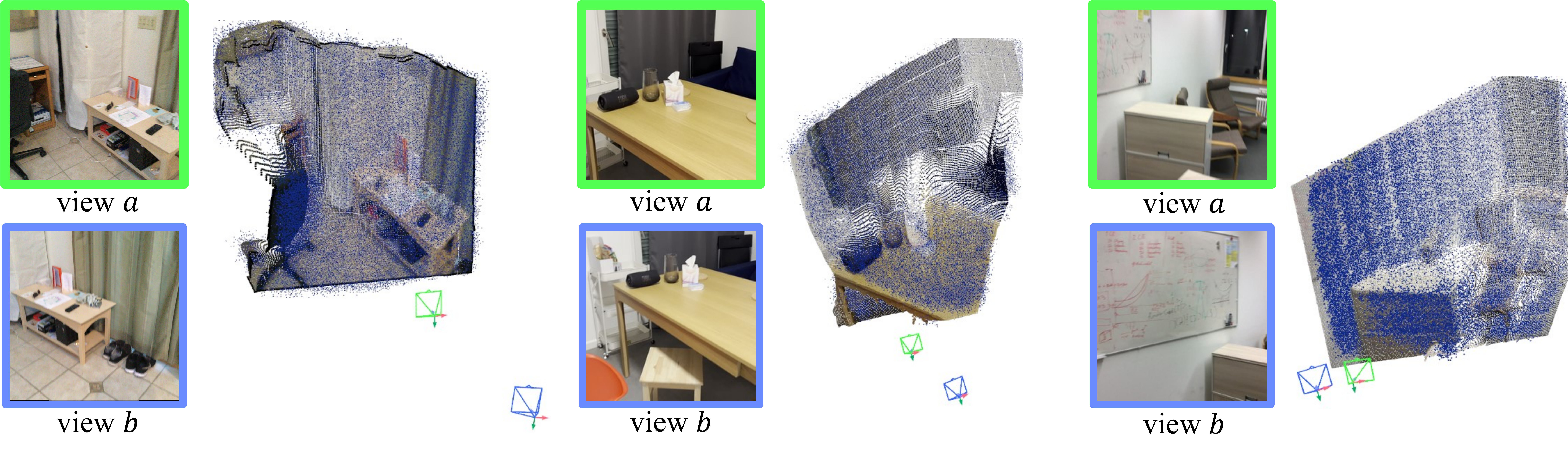}
\caption{
\textbf{Visualization of camera-conditioned geometry transformation.}
Given two input views, $a$ and $b$, we visualize the point map regressed from the geometry tokens $\bm{g}a$ of view $a$ as colored point clouds in the \green{coordinate frame of view $a$}.
We also visualize the point map regressed from the transformed tokens $\bm{\pi}^{a\rightarrow b}_{c}(\bm{g}_a)$, which is obtained by conditioning $\bm{g}_a$ on the camera token $c_{a\rightarrow b}$, as blue point clouds in the \textcolor{blue}{coordinate frame of view $b$}.
The two point clouds align closely under the pose transformation encoded in the camera token, demonstrating that the Camera-Conditioned Geometry Transform component of \ours operates as intended by transforming latent local geometry into the coordinate frames of other views.}
\label{fig:transfrom}

\vspace{-.2em}
\end{figure*}
\subsection{Component-Level Behavior}
In \cref{fig:transfrom}, we also visualize the camera-conditioned geometry transform component by comparing the geometry regressed directly from the local geometry tokens with the geometry regressed from the camera-conditioned transformed tokens. The two geometry predictions roughly align under the provided camera transform, which shows that the transform behaves as designed.

Together with the correspondence and residual visualizations already shown in the main text, it provides component-level evidence that the learned modules follow the intended BA-inspired behavior. Although we do not explicitly train each component to exactly match its BA counterpart, the end-to-end training results show that the learned modules still roughly play the intended functional roles. This further supports the view that the BA-inspired structural design is valuable.


\section{Multiview Extension Details}
To extend \ours to the multi-view setting, we generalize the two-view formulation (\cref{algo:two-view}) to operate on multiview inputs. 
Additionally, it requires a view graph $\mathcal{G}(\mathcal V,\mathcal E)$ as input, where each node $i\in\mathcal V$ represents a view with geometry tokens $\bm g_i$, and each edge $(i,j)\in\mathcal E$ is associated with a relative camera token $c_{i\rightarrow j}$. The overall architecture of \ours remains unchanged; the same refinement layer is applied without any modification to the network structure. The only difference lies in how tokens interact according to the view graph connectivity, shown in \cref{algo:multi-view}.

For each edge $(i,j)$, residual tokens $\bm r_{i\rightarrow j}$ are computed using the same procedure as in the two-view case: geometry tokens from view $i$ are first transformed into view $j$ using the camera token $c_{i\rightarrow j}$, latent correspondences are then obtained, and residuals are computed in latent space. Camera tokens are subsequently updated independently for each edge using cross-attention between the camera token and its associated residual tokens.

The only difference from the two-view formulation arises in the local geometry update. Instead of performing cross-attention between a pair of views, each view attends to geometry tokens from all neighboring views in the graph, transformed into the same coordinate frame using the updated camera tokens. This effectively changes the interaction from a one-view-to-one-view cross-attention to a one-view-to-many-views cross-attention, allowing each view to aggregate multiview geometric constraints.

Importantly, this extension does not modify the network architecture: the same refinement layer and parameters are reused without introducing additional modules or increasing the model depth or width. The multiview capability therefore emerges purely from the changed token connectivity defined by the input view-graph.

\noindent
\resizebox{0.48\textwidth}{!}{%
\begin{minipage}{0.6\textwidth}
\begin{algorithm}[H]
    \caption{Two-view \ours}
    \label{algo:two-view}
    \KwIn{$\bm g_a^{(k)}$, $\bm g_b^{(k)}$, $c_{a\rightarrow b}^{(k)}$}
    \vspace{2pt}
    \KwOut{$\bm g_a^{(k+1)}$, $\bm g_b^{(k+1)}$, $c_{a\rightarrow b}^{(k+1)}$}

    \vspace{3pt}
    {// Residual computation}\\
    $\tilde{\bm g}_a^{(k)} \leftarrow 
    \pi_{c^{(k)}}^{a\rightarrow b}(\bm g_a^{(k)})$
    
    $\bm g_b^{q,(k)} \leftarrow 
    \operatorname{LatentMatch}(\tilde{\bm g}_a^{(k)},\bm g_b^{(k)})$ 
    
    $\bm r_{a\rightarrow b}^{(k)} \leftarrow
    \operatorname{MLP}\!\left(
    f(\bm g_b^{q,(k)}) - f(\tilde{\bm g}_a^{(k)})
    \right)$
    
    \vspace{3pt}
    {// Camera update}\\
    $\Delta c_{a\rightarrow b}^{(k)} \leftarrow
    \operatorname{CrossAttn}\!\left(
    c_{a\rightarrow b}^{(k)},\;
    \bm r_{a\rightarrow b}^{(k)}
    \right)$
    
    $c_{a\rightarrow b}^{(k+1)} \leftarrow
    c_{a\rightarrow b}^{(k)} + \Delta c_{a\rightarrow b}^{(k)}$
    
    \vspace{5pt}
    {// Geometry update}\\
    $\Delta \bm g_a^{(k)} \leftarrow
    \operatorname{CrossAttn}\!\left(
    \bm g_a^{(k)},\;
    \bm \pi_{c^{(k+1)}}^{b\rightarrow a}(\bm g_b^{(k)})
    \right)$
    
    $\Delta \bm g_b^{(k)} \leftarrow
    \operatorname{CrossAttn}\!\left(
    \bm g_b^{(k)},\;
    \bm \pi_{c^{(k+1)}}^{a\rightarrow b}(\bm g_a^{(k)})
    \right)$
    
    $\bm g_a^{(k+1)} \leftarrow \bm g_a^{(k)} + \Delta \bm g_a^{(k)}$
    
    $\bm g_b^{(k+1)} \leftarrow \bm g_b^{(k)} + \Delta \bm g_b^{(k)}$
\end{algorithm}

\end{minipage}
}\hfill
\resizebox{0.5\textwidth}{!}{%
\begin{minipage}{0.65\textwidth}
\begin{algorithm}[H]
    \caption{Multiview \ours}
    \label{algo:multi-view}
\KwIn{$\{\bm g_i^{(k)}\}_{i\in\mathcal V}$, $\{c_{i\rightarrow j}^{(k)}\}_{(i,j)\in\mathcal E}, \mathcal{G}(\mathcal{V},\mathcal{E})$}
\vspace{3pt}
\KwOut{$\{\bm g_i^{(k+1)}\}_{i\in\mathcal V}$, $\{c_{i\rightarrow j}^{(k+1)}\}_{(i,j)\in\mathcal E}$}

\vspace{3pt}
    {// Residual computation for each edge}\\
    \ForEach{$(i,j)\in\mathcal E$}{
    $\tilde{\bm g}_i^{(k)} \leftarrow 
    \pi_{c_{i\rightarrow j}^{(k)}}^{i\rightarrow j}(\bm g_i^{(k)})$ 
    
    $\bm g_j^{q,(k)} \leftarrow 
    \operatorname{LatentMatch}(\tilde{\bm g}_i^{(k)}, \bm g_j^{(k)})$ 
    
    $\bm r_{i\rightarrow j}^{(k)} \leftarrow 
    \operatorname{MLP}\!\left(
    f(\bm g_j^{q,(k)}) - f(\tilde{\bm g}_i^{(k)})
    \right)$
    }
    
    \vspace{3pt}
    {// Camera update for each edge}\\
    \ForEach{$(i,j)\in\mathcal E$}{
    $\Delta c_{i\rightarrow j}^{(k)} \leftarrow
    \operatorname{CrossAttn}\!\left(
    c_{i\rightarrow j}^{(k)},\;
    \bm r_{i\rightarrow j}^{(k)}
    \right)$
    
    $c_{i\rightarrow j}^{(k+1)} \leftarrow
    c_{i\rightarrow j}^{(k)} + \Delta c_{i\rightarrow j}^{(k)}$
    }
    
    \vspace{3pt}
    {// Geometry update for each view}\\
    \ForEach{$i\in\mathcal V$}{
    $\Delta \bm g_i^{(k)} \leftarrow 
    \operatorname{CrossAttn}\!\left(
    \bm g_i^{(k)},\;
    \{\bm \pi_{c_{j\rightarrow i}^{(k+1)}}^{j\rightarrow i}(\bm g_j^{(k)})\}_{j:(j,i)\in\mathcal E}
    \right)$
    
    $\bm g_i^{(k+1)} \leftarrow \bm g_i^{(k)} + \Delta \bm g_i^{(k)}$
}
    
\end{algorithm}
\end{minipage}
}

\begin{table}[ht]
\centering
\caption{\textbf{Quantitative comparison of 4 view inputs on BundleFusion~\cite{dai2017bundlefusion}.} \ours achieves comparable accuracy to prior methods while requiring significantly lower memory and substantially faster inference. }
\vspace{0.2em}
\resizebox{0.9\textwidth}{!}{%
\setlength{\tabcolsep}{6pt}
\renewcommand{\arraystretch}{1.1}
\begin{tabular}{lcccccc}
\toprule
Method 
& Trans. (m)
& Rot.  ($^\circ$)
& Rel. 
& $\delta<1.25$ 
& Time (ms) 
& Peak Mem (GB) \\
\midrule
MapAny~\cite{keetha2026mapanything} & 0.14 & 29.65 & 0.36 & 0.30 & \underline{62.95} & \underline{3.42} \\
VGGT~\cite{wang2025vggt}   & \bf 0.06 & \bf 10.72 & \bf 0.15 & \bf 0.79 & 230.13 & 5.11 \\
\textbf{BA-T}   & \underline{0.09} & \underline{16.39} & \underline{0.20} & \underline{0.70} & \bf 50.62 & \bf 1.35 \\
\bottomrule
\end{tabular}
}
\label{tab:4view}
\end{table}

\begin{figure*}[!ht]
\centering
\includegraphics[width=0.95\linewidth]{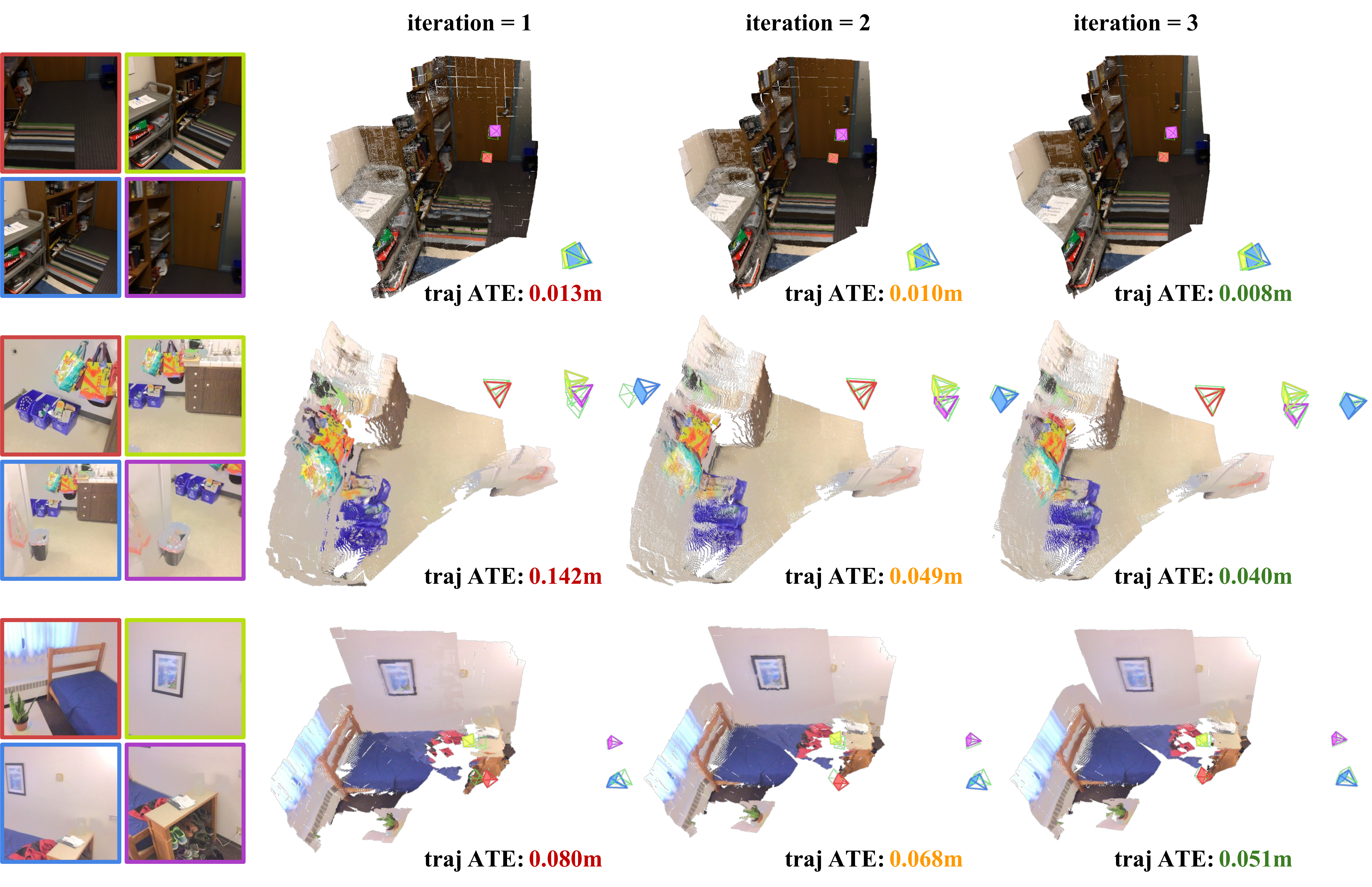}
\caption{
\textbf{More qualitative 4-view results.}
}
\label{fig:more_mv}
\end{figure*}

For multi-view setting, we train an additional 12 hours on 4-view inputs.
We evaluate \ours on 4-view inputs from BundleFusion~\cite{dai2017bundlefusion} (547 4-view samples) and compare it with recent multi-view methods~\cite{wang2025vggt,keetha2026mapanything}. In this experiment, the four views form a fully connected view graph, where a camera token is defined for every pair of views. The evaluation metrics include absolute trajectory error (Trans.), absolute rotation error (Rot.), median relative depth error (Rel.), $\delta<1.25$ accuracy, inference time, and peak memory.
As shown in \cref{tab:4view}, \ours achieves accuracy comparable to prior methods while being substantially more efficient, using only 39\% of the peak memory of MapAnything. These results demonstrate that \ours scales naturally to the multi-view setting while maintaining strong accuracy and significantly improving computational efficiency. Additional multi-view results are presented in \cref{fig:more_mv}.

\section{Limitations and Failure Modes}
\label{sec:limitations}
The failure modes of the current model follow directly from its training distribution. Ground-truth geometry is far easier to obtain for indoor, static scenes, so our training mixture is indoor and static and uses pinhole or near-pinhole cameras throughout. Consequently we \emph{cannot} claim generalisation to (i) \textbf{dynamic scenes}, (ii) \textbf{outdoor scenes}, in particular large-scale outdoor scenes, or (iii) \textbf{heavily distorted inputs} such as strong fisheye or wide-angle lenses. This is a limitation of the current \emph{model} rather than of the architecture: the BA-inspired refinement layer makes no assumption that the scene is static, indoor or undistorted, and we expect these regimes to be addressable by broadening the training data.

\begin{table}[htb]
\centering
\small
\begin{tabular}{ll}
\hline
Dataset & License / Terms of Use \\
\hline
ScanNet~\cite{dai2017scannet} & Non-commercial \\
ScanNet++~\cite{yeshwanthliu2023scannetpp} & Non-commercial \\
ARKitScenes~\cite{dehghan2021arkitscenes} & Non-commercial \\
CO3D~\cite{reizenstein21co3d} & CC BY-NC 4.0 \\
ASE~\cite{avetisyan2024ase} & Non-commercial \\
Replica~\cite{replica19arxiv} & Non-commercial \\
\hline
\end{tabular}
\vspace{4pt}
\caption{Summary of the licenses or terms of use for the datasets used in training.}
\label{tab:dataset_licenses}
\end{table}

\clearpage


\newpage
\section*{NeurIPS Paper Checklist}

\begin{enumerate}

\item {\bf Claims}
    \item[] Question: Do the main claims made in the abstract and introduction accurately reflect the paper's contributions and scope?
    \item[] Answer: \answerYes{}
    \item[] Justification: The abstract and introduction accurately describe the main contribution as a BA-inspired iterative Transformer in latent token space, and the experimental sections support the claimed gains in pose, geometry consistency, and efficiency.

\item {\bf Limitations}
    \item[] Question: Does the paper discuss the limitations of the work performed by the authors?
    \item[] Answer: \answerYes{}
    \item[] Justification: The paper discusses key limitations, including the current two-view implementation, the limited exploration of broader multiview settings.

\item {\bf Theory assumptions and proofs}
    \item[] Question: For each theoretical result, does the paper provide the full set of assumptions and a complete (and correct) proof?
    \item[] Answer: \answerNA{}
    \item[] Justification: The paper does not present theoretical results or proofs.

    \item {\bf Experimental result reproducibility}
    \item[] Question: Does the paper fully disclose all the information needed to reproduce the main experimental results of the paper to the extent that it affects the main claims and/or conclusions of the paper (regardless of whether the code and data are provided or not)?
    \item[] Answer: \answerYes{}
    \item[] Justification: The main paper and supplement disclose the experimental settings needed to reproduce the reported results.

\item {\bf Open access to data and code}
    \item[] Question: Does the paper provide open access to the data and code, with sufficient instructions to faithfully reproduce the main experimental results, as described in supplemental material?
    \item[] Answer: \answerNo{} 
    \item[] Justification: The data used is publicly available, and the code will be released upon publication with instructions for reproducing the main results.

\item {\bf Experimental setting/details}
    \item[] Question: Does the paper specify all the training and test details (e.g., data splits, hyperparameters, how they were chosen, type of optimizer) necessary to understand the results?
    \item[] Answer: \answerYes{}
    \item[] Justification: Training datasets, settings and hardware are all described in the main paper and supplement.

\item {\bf Experiment statistical significance}
    \item[] Question: Does the paper report error bars suitably and correctly defined or other appropriate information about the statistical significance of the experiments?
    \item[] Answer: \answerNo{}
    \item[] Justification: The paper reports main results without error bars or significance tests, following the standard practice in the field.

\item {\bf Experiments compute resources}
    \item[] Question: For each experiment, does the paper provide sufficient information on the computer resources (type of compute workers, memory, time of execution) needed to reproduce the experiments?
    \item[] Answer: \answerYes{}
    \item[] Justification: The paper specifies training setting and resources and reports inference memory and runtime statistics for the main experiments in supplement.
    
\item {\bf Code of ethics}
    \item[] Question: Does the research conducted in the paper conform, in every respect, with the NeurIPS Code of Ethics \url{https://neurips.cc/public/EthicsGuidelines}?
    \item[] Answer: \answerYes{}
    \item[] Justification: The research is a standard reconstruction method evaluated on public datasets, and we are not aware of any ethical issues that would violate the NeurIPS Code of Ethics.

\item {\bf Broader impacts}
    \item[] Question: Does the paper discuss both potential positive societal impacts and negative societal impacts of the work performed?
    \item[] Answer: \answerNo{}
    \item[] Justification: The paper does not discuss broader impacts explicitly. The method has clear positive applications in reconstruction, but potential misuse or downstream harms are not addressed in the text.
    
\item {\bf Safeguards}
    \item[] Question: Does the paper describe safeguards that have been put in place for responsible release of data or models that have a high risk for misuse (e.g., pre-trained language models, image generators, or scraped datasets)?
    \item[] Answer: \answerNA{}
    \item[] Justification: The paper does not release a high-risk model or dataset that requires special safeguards.

\item {\bf Licenses for existing assets}
    \item[] Question: Are the creators or original owners of assets (e.g., code, data, models), used in the paper, properly credited and are the license and terms of use explicitly mentioned and properly respected?
    \item[] Answer: \answerYes{}
    \item[] Justification: The paper cites the datasets and baseline methods used, the license and terms of use for the datasets are mentioned in the appdendix.

\item {\bf New assets}
    \item[] Question: Are new assets introduced in the paper well documented and is the documentation provided alongside the assets?
    \item[] Answer: \answerNA{}
    \item[] Justification: The paper does not introduce a newly released dataset or other standalone asset package.

\item {\bf Crowdsourcing and research with human subjects}
    \item[] Question: For crowdsourcing experiments and research with human subjects, does the paper include the full text of instructions given to participants and screenshots, if applicable, as well as details about compensation (if any)? 
    \item[] Answer: \answerNA{}
    \item[] Justification: The paper does not involve crowdsourcing or human-subject experiments.

\item {\bf Institutional review board (IRB) approvals or equivalent for research with human subjects}
    \item[] Question: Does the paper describe potential risks incurred by study participants, whether such risks were disclosed to the subjects, and whether Institutional Review Board (IRB) approvals (or an equivalent approval/review based on the requirements of your country or institution) were obtained?
    \item[] Answer: \answerNA{}
    \item[] Justification: The paper does not involve research with human subjects.

\item {\bf Declaration of LLM usage}
    \item[] Question: Does the paper describe the usage of LLMs if it is an important, original, or non-standard component of the core methods in this research? Note that if the LLM is used only for writing, editing, or formatting purposes and does \emph{not} impact the core methodology, scientific rigor, or originality of the research, declaration is not required.
    \item[] Answer: \answerNA{}
    \item[] Justification: LLMs are not part of the core method; they are not used as an important, original, or non-standard component of the research.

\end{enumerate}

\end{document}